\PassOptionsToPackage{utf8}{inputenc}
\documentclass[twocolumn]{article}
\RequirePackage{natbib,inputenc,graphicx,amsmath,array,color,amssymb,amsthm,times,crop,flushend,stfloats,chngpage}
\usepackage{booktabs}
\usepackage{rotating}
\usepackage[hidelinks]{hyperref}
\usepackage{multirow}
\usepackage{nameref}
\usepackage{cleveref}
\usepackage{amsfonts}
\usepackage{ragged2e}
\usepackage[american]{babel}

\usepackage[sectionbib]{chapterbib}
\usepackage{caption}
\begin{document}

\title{Insights Into the Inner Workings of Transformer Models \\ for Protein Function Prediction\footnote{Published originally in \textit{Bioinformatics} by Oxford University Press under the \href{https://creativecommons.org/licenses/by/4.0/}{CC BY 4.0} license. \copyright~The Authors 2024. Wenzel, M., Grüner, E. and Strodthoff, S. (2024). Insights into the inner workings of transformer models for protein function prediction, \textit{Bioinformatics}, btae031. \url{https://doi.org/10.1093/bioinformatics/btae031}}}
\author{Markus Wenzel$^1$ \and Erik Gr\"uner$^1$ \and Nils Strodthoff$^2$}
\date{%
    \small{\textit{$^1$ Department of Artificial Intelligence, Fraunhofer Institute for Telecommunications, Heinrich-Hertz-Institut, HHI, \\ Einsteinufer 37, 10587 Berlin, Germany}}\\%
    \small{\textit{$^2$ School VI - Medicine and Health Services, Carl von Ossietzky University of Oldenburg, \\ Ammerl\"ander Heerstr. 114-118, 26129 Oldenburg, Germany}}\\[1ex]%
    markus.wenzel@hhi.fraunhofer.de, nils.strodthoff@uol.de
    }
\maketitle

\abstract{
\textbf{Motivation:} 
We explored how explainable artificial intelligence (XAI) can help to shed light into the inner workings of neural networks for protein function prediction, by extending the widely used XAI method of integrated gradients such that latent representations inside of transformer models, which were finetuned to Gene Ontology term and Enzyme Commission number prediction, can be inspected too.
\textbf{Results:} The approach enabled us to identify amino acids in the sequences that the transformers pay particular attention to, and to show that these relevant sequence parts reflect expectations from biology and chemistry, both in the embedding layer and inside of the model, where we identified transformer heads with a statistically significant correspondence of attribution maps with ground truth sequence annotations (e.g. transmembrane regions, active sites) across many proteins.
\textbf{Availability and Implementation:} Source code can be accessed at \href{https://github.com/markuswenzel/xai-proteins}{this https URL}.
}

\section{Introduction}

\subsection{Protein function prediction}

\textbf{Proteins -- constituents of life}. Proteins are versatile molecular machines, performing various tasks in basically all cells of every organism, and are modularly constructed from chains of amino acids. Inferring the function of a given protein merely from its amino acid sequence is a particularly interesting problem in bioinformatics research. 

Function prediction can help to rapidly provide valuable pointers in the face of so far unfamiliar proteins of understudied species, such as of emerging pathogens. 
Moreover, it makes the analysis of large, unlabeled protein data sets possible, which becomes more and more relevant against the backdrop of the massive and evermore growing databases of unlabeled nucleic acid sequences, which again can be translated into amino acid sequences. Next-generation DNA sequencers can read the nucleic acid sequences present in a sample or specimen at decreasing costs \citep{mardis2017dna, shendure2017dna}, much faster than experimenters can determine the function of the genes and corresponding proteins. Therefore, databases with genes and corresponding amino acid sequences grow much more rapidly than those of respective experimental gene and protein labels or annotations.
Besides, gaining knowledge about the mapping between amino acid sequence and protein function can help to engineer proteins for dedicated purposes too \citep{alley2019unified, yang2019machine, ferruz2022protgpt2, Hie2022, madani2023large}.

\textbf{Machine learning approaches to protein function prediction} can include inferring enzymatic function \citep{dalkiran2018ecpred, li2018deepre, zou2019mldeepre, yu2023enzyme}, Gene Ontology (GO) terms \citep{kulmanov2017deepgo, You2018Golabeler, you2018deeptext2go, you2021deepgraphgo, kulmanov2019deepgoplus, kulmanov2022deepgozero, strodthoff2020udsmprot, littmann2021embeddings}, protein--protein/--drug interaction, remote homology, stability, sub-cellular location, and other properties \citep{Rao2019, Bepler2021}. For structure prediction, the objective is to infer how the amino acid sequence folds into the secondary \citep{zhang2018prediction, Rives2021} and tertiary protein structure \citep{Torrisi2020, AlQuraishi2021, jumper2021highly, Weissenow2022}. 
Several of the prediction tasks can be approached as well by transferring labels from similar sequences obtained via multiple sequence alignment (MSA) \citep{buchfink2014fast, gong2016gofdr}.
Protein prediction models are compared by the scientific community in systematic performance benchmarks, e.g. for function annotation \citep[CAFA, ][]{radivojac2013large, jiang2016expanded, zhou2019cafa}, for structure prediction \citep[CASP, ][]{kryshtafovych2019critical, kryshtafovych2021critical}, or for several semi-supervised tasks \citep{Rao2019, fenoy2022transfer}.
Machine learning methods are continuing to win ground in comparison to MSA-techniques with respect to performance, have a short inference time, and can process sequences from the so-called ``dark proteome'' too, where alignments are not possible \citep{perdigao2015unexpected, Rao2019, lin2023evolutionary}.

\subsection{Protein language modeling and transfer learning}

\textbf{Relations to NLP}. Amino acid sequences share some similarities with the sequences of letters and words occurring in written language, in particular with respect to the complex interrelationships between distant elements, which are arranged in one-dimensional chains. Thus, recent progress in research on natural language processing (NLP) employing language modeling in a transfer learning scheme \citep{howard2018universal} has driven forward protein function prediction too \citep[e.g.~][]{strodthoff2020udsmprot}.

\textbf{Self-supervised pretraining}. Typically, a language model is first pretrained on large numbers of unlabeled sequences in an unsupervised fashion, e.g. by learning to predict masked tokens (cloze task) or the respective next token in the sequences (which is why this unsupervised approach is also dubbed self-supervised learning). In this way, the model learns useful representations of the sequence statistics (i.e. language). These statistics possibly arise because the amino acid chains need to be stable under physiological conditions and are subject to evolutionary pressure. The learned representations can be transferred to separate downstream tasks, where the pretrained model can be further finetuned in a supervised fashion on labeled data, which are usually available in smaller amounts, considering that sequence labeling by experimenters is costly and lengthy. 

\textbf{Model architectures}. Transformer models \citep{vaswani2017attention} making use of the attention mechanism \citep{niu2021review}, such as bidirectional encoder representations from transformers \citep[BERT, ][]{devlin2018bert} are currently prevailing architectures in NLP. Transformers have been recently applied to the study of amino acid sequences too, pushing the state of the art in the field of proteomics as well \citep{Rao2019, Rao2021, Nambiar2020, Bepler2021, littmann2021embeddings, Rives2021, brandes2022proteinbert, Elnaggar2022, fenoy2022transfer, unsal2022learning, lin2023evolutionary, olenyi2023}.
Recurrent neural networks (RNNs) using long short term memory (LSTM) cells are another model architecture that is particularly suited to process sequential data. RNNs have been successfully employed to protein \citep{strodthoff2020udsmprot} and peptide \citep{vielhaben2020usmpep} property prediction as well, within the scheme of language modeling combined with transfer learning, as sketched out above.

\subsection{Explainable machine learning}

\textbf{Need for explainability}. Transformers and other modern deep learning models are notorious for having often millions and sometimes billions of trainable parameters, and it can be very difficult to interpret the decision making logic or strategy of such complex models. The research field of explainable machine learning \citep{lundberg2017unified, montavon2018methods, arrieta2020explainable, tjoa2020survey, covert2021explaining, samek2021explaining} aims at developing methods that enable humans to better interpret -- or to a limited degree: understand -- such ``opaque'', complex models. In certain cases, it was demonstrated that the methods can even help to uncover flaws and unintended biases of the models, such as being mislead by spurious correlations in the data \citep{lapuschkin2019unmasking}. 

\textbf{Attribution methods}, such as integrated gradients (IG) \citep{sundararajan2017}, layerwise-relevance propagation \citep{bach2015pixel, binder2016lrp} or gradient-weighted class activation mapping \citep{selvaraju2017grad}, make it possible to identify those features in the input space that the model apparently focuses on, because these features turn out to be particular relevant for the final classification decision of the model. Further examples of model explainability methods include probing classifiers \citep{belinkov2022probing}, testing with concept activation vectors \citep{kim2018interpretability}, and studying the attention mechanism \citep{jain2019, serrano2019, bai2021, niu2021review}. Explainability methods have been employed in NLP too \citep{Arras19eval, manning2020emergent, chefer2021transformer, pascual21eval}. Moreover, researchers have started to explore using explainability methods in the area of protein function prediction \citep{upmeier2019leveraging, taujale2021mapping, vig2021bertology, hou2023learning, vu2023linguistically, zhou2023phosformer}.

\subsection{Contributions of the article}

\begin{figure}
\centering
\includegraphics[width=\columnwidth]{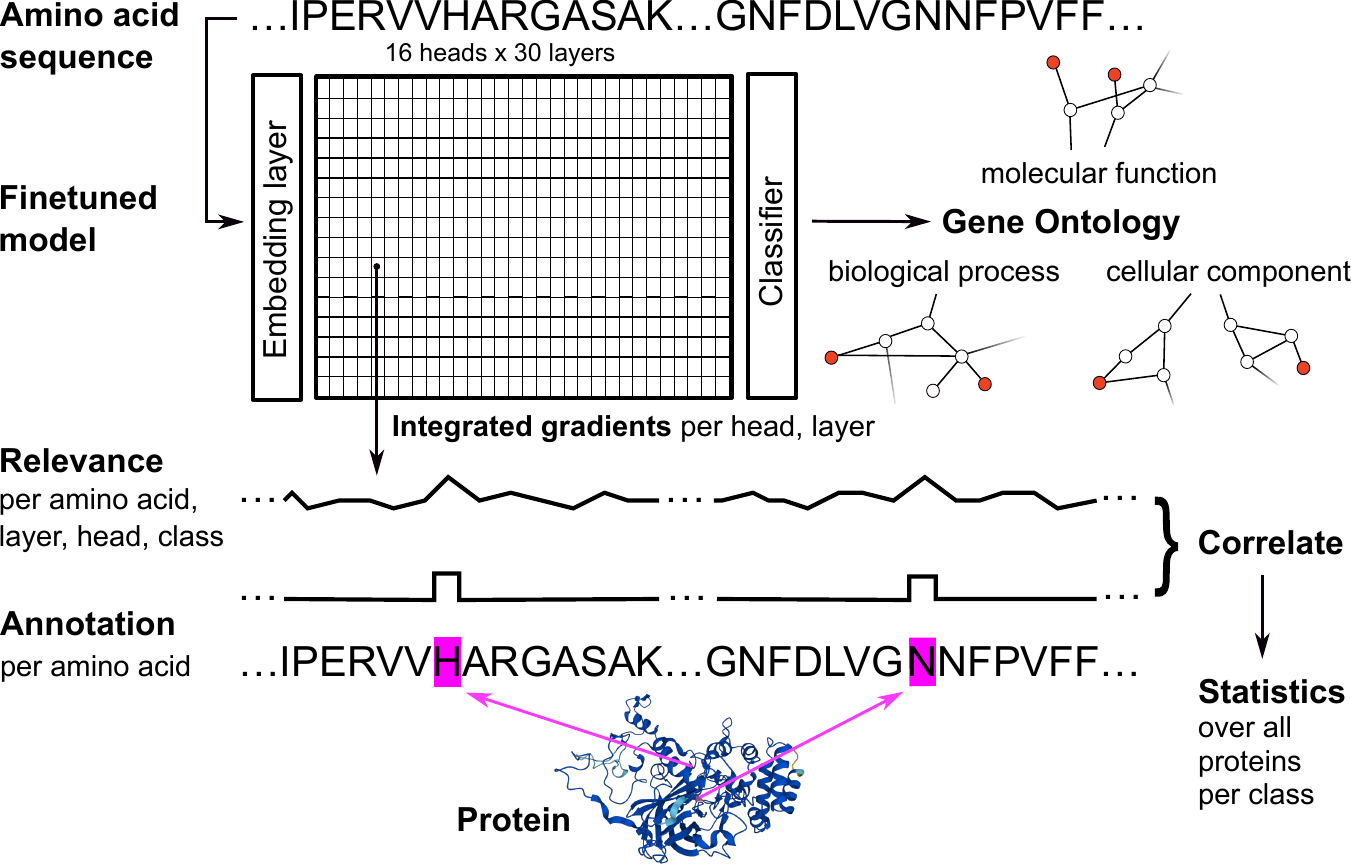}
\caption{Illustration of the experimental design.
Top: From the amino acid sequence, the finetuned transformer model infers the applicable Gene Ontology (GO) terms (represented as multi-label class membership vector).
(The depicted exemplary ``catalase-3'' should be labeled with the GO terms ``catalase activity'' as ``molecular function'', ``response to hydrogen peroxide'' as ``biological process'', ``cytoplasm'' as ``cellular component'' etc.; about 5K of about 45K GO terms were considered.)
Center: Relevance indicative for a selected GO term was attributed to the amino acids per protein and correlated with corresponding annotations per amino acid. This correlation between relevance attributions and annotations was then statistically assessed across the test data set proteins. The analysis was conducted for the embedding layer and ``inside'' of the model, for each head in each layer, and was repeated for different GO terms (see \Cref{subsec:revealing}).
Bottom: Specific amino acids of a protein are annotated in sequence databases like UniProt, because they serve as binding or active sites or are located in the cell membrane etc.
Active sites can, e.g. be found at the histidine (``H'' at position 65) and asparagine (``N'' at position 138) of ``\href{https://alphafold.ebi.ac.uk/entry/O48560}{catalase-3}'' (protein structure prediction created by \href{https://alphafold.com/}{AlphaFold} -- ``AlphaFold Data Copyright (2022) DeepMind Technologies Limited'' -- under the \href{https://creativecommons.org/licenses/by/4.0/}{CC-BY 4.0 licence}  \citep{jumper2021highly, varadi2021alphafold}).}
\label{fig:experimental_design}
\end{figure}

\textbf{Goal of the study}. Building upon this previous research on the interpretation of protein classification models, we aimed at exploring how explainability methods can further help to gain insights into the inner workings of the now often huge neural networks, and proceeded as follows.

\textbf{Specific contributions}. First, we finetuned pretrained transformers on selected prediction tasks and could push or reach the state-of-the-art (see Supplementary \Cref{appendix:results_and_discussion}). Then, we quantified the relevance of each amino acid of a protein for the function prediction model. 
Subsequently, we investigated if these relevant sequence regions match expectations informed by knowledge from biology or chemistry, by correlating the relevance attributions with annotations from sequence databases (see \Cref{fig:experimental_design}).
For instance, we addressed the question if a classification model that is able to infer if a protein is situated in the cell membrane does indeed focus systematically on transmembrane regions or not.
We conducted this analysis on the embedding level and ``inside'' of the model with a novel adaptation of IG. In this way, we identified transformer heads with a statistically significant correspondence of the attribution maps with ground truth annotations, across many proteins and thus going beyond anecdotes of few selected cases.

\enlargethispage{12pt}

\section{System and methods}

\subsection{Revealing insights into function prediction models}
\label{subsec:revealing}

\textbf{The prediction tasks} of inferring GO terms and Enzyme Commission (EC) numbers, that the proteins are labeled with, from their amino acid sequence are detailed in Supplementary \Cref{appendix:system_and_methods}. This Supplementary material also explains the finetuning of the transformers ``ProtBert-BFD'' and ``ProtT5-XL-UniRef50'' \citep{Elnaggar2022} and ``ESM-2'' \citep{lin2023evolutionary} on the GO and EC tasks, and contains statements about data availability and composition.

\textbf{Overall approach}. We investigated whether specific positions or areas on the amino acid sequence that had been annotated in sequence data bases are particularly relevant for the classification decision of the model (see \Cref{fig:experimental_design}). 
Annotations included UniProtKB/Swiss-Prot ``active'' and ``binding sites'', ``transmembrane regions'', ``short sequence motifs'', and PROSITE patterns related to a GO term and its children terms in the ontology. Definitions of the aforementioned UniProt annotations (per amino acid) and matching GO terms (class labels of proteins) are compiled in Supplementary \Cref{tab:uniprot_go} (tables/figures with prefix letters are shown in the Supplementary material).
First, we attributed relevance indicative for a given class (either a selected GO term or EC number) to each amino acid of a protein. Then, we correlated the relevance heat map obtained for the amino acid chain of a protein with corresponding binary sequence annotations. 
To study the information representation within the model, the explainability analysis was conducted at the embedding layer and repeated ``inside'' of the model, separately for its different heads and layers, using a novel method building upon IG, described below in \Cref{sec:algorithm}.

\textbf{Experimental setup}. For the experimental evaluation, we focus on the pretrained ProtBert model that was finetuned either to the multi-label GO-classification on the GO ``2016'' data set, or to the multi-class EC number classification on the ``EC50 level L1'' data set.
We consider the comparatively narrow EC task in addition to the much more comprehensive GO prediction, because the test split of the EC data set contains a larger number of samples that are both labeled per protein and annotated per amino acid, which is beneficial for the conducted explainability analysis.
We observed that larger models tend to perform numerically better than smaller models (see Supplementary \Cref{appendix:results_and_discussion}). Given our focus on methodological matters of model interpretation, we deliberately studied ProtBert (420M parameters), because it is better manageable, due to its considerably smaller memory footprint, in comparison to the larger ProtT5 (1.2B).

\section{Algorithm}
\label{sec:algorithm}

\textbf{Integrated gradients} \citep{sundararajan2017} represents a model-agnostic attribution method, which can be characterized as unique attribution method satisfying a set of four axioms (Invariance, Sensitivity, Linearity, and Completeness). In this formalism, the attribution for feature $i$ is defined via the line integral (along a path, parameterized as $\gamma(t)$ with $t\in[0,1]$, between some chosen baseline $\gamma(0)=x'$ and the sample to be explained $\gamma(1)=x$),
\begin{equation}
\text{IG}^\gamma_i = \int_0^1 \text{d}\alpha \frac{\partial F(\gamma(\alpha))}{\partial \gamma_i} \frac{\text{d}\gamma_i}{\text{d}\alpha}\,,
\label{eq:integrated_gradients}
\end{equation}
where $F$ is the function we aim to explain. Choosing $\gamma$ as straight line connecting $x'$ and $x$ makes IG the unique method satisfying the four axioms from above and an additional symmetry axiom. This path is the typical choice in applications applied directly to the input layer for computer vision or to the embedding layer for NLP. The approach can be generalized to arbitrary layers if one replaces $x$ and $x'$ by the hidden feature representation of the network up to this layer (referred to as ``layer IG'' in the popular ``Captum'' library \citep{captum}).

\begin{figure}
\centering
\includegraphics[width=\columnwidth]{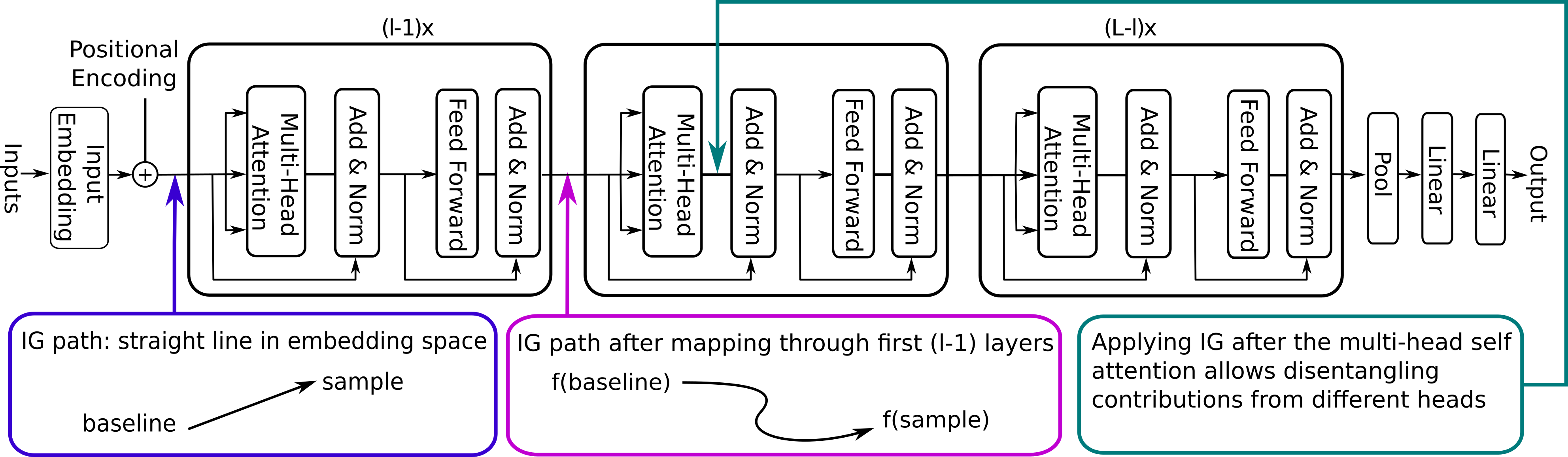}
\caption{Visualization of the explainability method based on IG that can attribute relevance to sequence tokens (here: amino acids) separately for each head and layer of the transformer \citep[adapted from][]{vaswani2017attention}.}
\label{fig:integrated_gradients}
\end{figure}

\textbf{Head-specific attribution maps}. To obtain attributions for individual heads, we have to target the output of the multi-head self-attention (MHSA) block of a particular layer; see \Cref{fig:integrated_gradients} for a visualization of the transformer architecture. Properly separating the attributions of the individual heads from the attribution contribution obtained from the skip connection necessitates to target directly the output of the MHSA. Now, one cannot just simply choose an integration path that connects baseline and sample as encoded by the MHSA block because the input for the skip connection has to be varied consistently. To keep an identical path in all cases, we fix the integration path as a straight line in the embedding layer, which then gets encoded into a, in general, curvilinear path seen as input for some intermediate layer. Choosing not a straight path only leads to the violation of the symmetry axiom, which is not of paramount practical importance in this application; see \citep{ward2020exploration,kapishnikov2021guided} for other applications with IG applied to general paths.
For every sample, this application of IG yields a relevance map of shape $\text{seq}\times n_\text{model}$, where the first $n_\text{model}/n_\text{heads}$ entries in the last dimension correspond to the first head, followed by the second head etc. By summing over $n_\text{model}/n_\text{heads}$ entries in the last dimension, we can reduce the relevance map to a $\text{seq}\times n_\text{heads}$ attribution map, i.e. one relevance sequence per head. 

\textbf{Correlation coefficients and statistical significance}.
Each sequence of relevance attributions can then be correlated with sequence annotations to find out if the model focuses on the annotated amino acids. Coefficients of point biserial correlation \citep{kornbrot2005point}, which is equivalent to Pearson correlation, were calculated between the continuous relevance values and the corresponding binary annotations per amino acid. This correlation analysis was conducted separately for each head in each transformer layer.
The resulting correlation coefficients were then assembled into a $n_\text{layer}\times n_\text{head}$ matrix per protein, which entered the subsequent statistical analysis across proteins. Summary statistics over all proteins (which belong to the respective GO or EC class, and, which are part of the respective test data set split) were obtained by computing Wilcoxon signed-rank tests across the correlation coefficients. The resulting p-values were corrected for the multiple tests per condition (16 heads times 30 layers equals 480 hypothesis tests) by controlling the false discovery rate \citep{benjamini1995controlling}.

\textbf{Summed attribution maps}.
In parallel to the correlation analysis, we furthermore sum the aforementioned attribution map along the sequence dimension, and obtain $ n_\text{heads}$ entries that specify the relevance distribution onto the different heads. 
We can carry out the same procedure for every transformer layer and combine all results into a $n_\text{layer}\times n_\text{head}$ relevance map of summed attributions. This map makes it possible to identify heads with a positive relevance with respect to the selected class. One map was obtained per protein. Heads with a significantly positive relevance were singled out by calculating a summary statistic across proteins with the Wilcoxon signed-rank test.
Finally, the two parallel analysis tracks were combined by identifying transformer heads that feature both a significantly positive (A) relevance-annotation-correlation and (B) relevance (this overlay is displayed in the figures by masking A with B).

\section{Implementation}
\label{sec:implementation}

Supplementary \Cref{appendix:implementation} shows implementation details.

\enlargethispage{6pt}

\section{Results and Discussion}

\subsection{Predictive performance}

The performance results for the ProtT5, ProtBert, and ESM-2 transformers finetuned to the GO and EC protein function tasks are presented in Supplementary \Cref{tab:go_cafa3,tab:go_2016,tab:ec_exp} in Supplementary\Cref{appendix:results_and_discussion}. In summary, we show that finetuning pretrained large transformer models leads to competitive results, in particular in the most relevant comparison in the single-model category, often on par with MSA-approaches.
Larger models lead the rankings, with ProtT5 competing with ESM-2.
Finetuning the entire model including the encoder shows its particular strength in the ``CAFA3'' benchmark.

\subsection{Explainability analysis: embedding layer}
\label{subsec:results_explainability_embedding}

\textbf{Research question}. Starting with embedding layer attribution maps, as the most widely considered type of attribution, we investigate whether there are significant correlations between attribution maps and sequence annotations from external sources (see \Cref{subsec:revealing}). We aim to answer this question in a statistical fashion going beyond anecdotal evidence based on single examples, which can sometimes be encountered in the literature.

\begin{figure}
\centering
\begin{tabular}{ccc}
\tiny{Membrane} & \tiny{Catalytic activity} & \tiny{Binding} \\
\includegraphics[height=0.34\columnwidth]{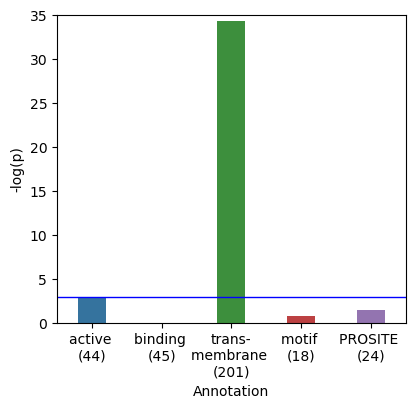} &
\includegraphics[trim={1.4cm 0 0 0}, clip, height=0.34\columnwidth]{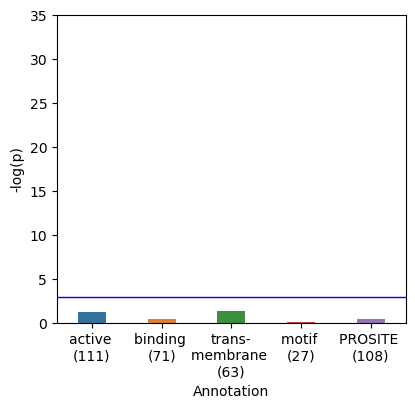} &
\includegraphics[trim={1.4cm 0 0 0}, clip, height=0.34\columnwidth]{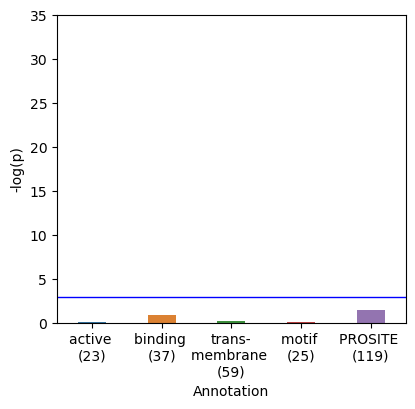}
\end{tabular}
\caption{Attribution maps for the embedding layer of ProtBert finetuned to GO term classification were correlated with sequence annotations. Relevance attributions indicative for the GO label ``membrane'' correlate significantly with UniProt annotations as ``transmembrane regions'' (p$<$0.05, i.e. above blue line). Attribution-annotation-correlation was not observed for GO ``catalytic activity'' and ``binding'. Numbers of test split samples both labeled with the GO term and annotated per amino acid are listed below the x-axis.}
\label{fig:go_embedding}
\end{figure}

\textbf{GO prediction: GO ``membrane'' attributions correlate in particular with UniProt ``transmembrane regions''}.
\Cref{fig:go_embedding} shows the results of the explainability analysis for the embedding layer of ProtBert finetuned to GO classification. The relevance of each amino acid indicative for selected GO terms was computed with IG, and then correlated with UniProt and PROSITE sequence annotations. Subsequently, it was tested whether the correlation coefficients across all annotated proteins from the test set were significantly positive (see \Cref{subsec:revealing}). A significant correlation was observed when relevance attributions indicative for the GO label ``membrane'' were correlated with UniProt ``transmembrane regions'' (p$\ll$0.05). 
Correlation was not observed in the GO ``catalytic activity'' and ``binding'' cases.

The pretrained model is expected to contain substantial information already prior to finetuning; e.g. \citet{bernhofer2022tmbed} had identified transmembrane regions using the pretrained ProtT5. Therefore, we inspected the GO membrane case in more detail. The pretrained but not finetuned ProtBert (combined with a classification head trained for the same number of epochs) resulted also in a significantly positive correlation of embedding level attributions to the GO term ``membrane'' with transmembrane regions only. Thus common patterns emerge between the pretrained and the finetuned ProtBert.

\begin{figure}
\centering
\includegraphics[height=0.4\columnwidth]{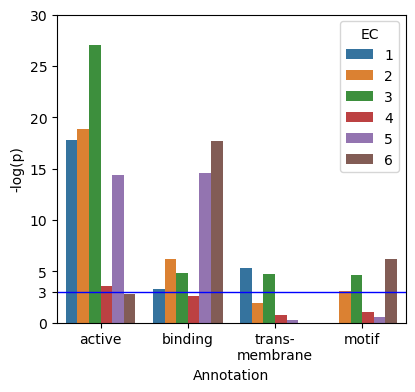}
\includegraphics[height=0.4\columnwidth]{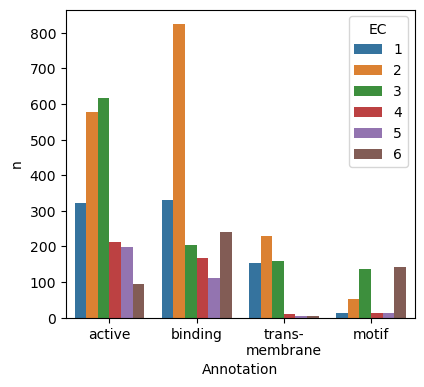}
\caption{Attribution maps calculated for the embedding layer of ProtBert finetuned to ``EC50 level~L1'' classification were correlated with UniProt sequence annotations. Left: Relevance attributions correlated significantly (p$<$0.05, i.e. above blue line) with ``active sites'' and ``binding sites'' for five out of six EC classes, and with ``transmembrane regions'' and ``short sequence motifs'' for two, respectively, three EC classes each. Right: Numbers of annotated samples in the test split per annotation type and EC class.}
\label{fig:ec_embedding}
\end{figure}

\textbf{EC prediction: attributions correlate significantly with several types of sequence annotations}.
\Cref{fig:ec_embedding} shows the results of the explainability analysis for the embedding layer of ProtBert finetuned to EC number classification (`EC50 level~L1'' data set; i.e., the differentiation between the six main enzyme classes). Relevance per amino acid for each of the six EC classes was correlated with the UniProt annotations as ``active sites'', ``binding sites'', ``transmembrane regions'', and ``short sequence motifs'. It can be observed that the relevance attributions correlated significantly (p$<$0.05) with ``active site'' and ``binding site'' annotations for five out of six EC classes, and with ``transmembrane regions'' and ``short sequence motifs'' for two, respectively, three EC classes. (Supplementary \Cref{fig:ec_embedding-EC4050L12} shows that positive relevance-annotation-correlation was observed for all annotation types for ``EC40'' and ``EC50'' on both levels ``L1'' and ``L2'' for several enzyme (sub-) classes.)

\textbf{Discussion}.
Attribution maps obtained for the embedding layer correlated with UniProt annotations on the amino acid level, in particular, in the EC case, but also for the GO term ``membrane'. To summarize, across two tasks, we provide first quantitative evidence for the meaningfulness and specificity of attribution maps beyond anecdotal evidence. Note that the EC case has the benefit of often several hundred annotated samples contained in the test split (except for ``transmembrane regions'' and ``motifs'; see right panel of \Cref{fig:ec_embedding}). In comparison, the GO case provides fewer samples in the test split of the data set that were also annotated on the amino acid level (see numbers in brackets below the x-axis in \Cref{fig:go_embedding}).

\subsection{Explainability analysis: peeking inside the transformer}
\label{subsec:explainability_analysis_peeking}

\textbf{Research question}.
Given the encouraging results presented in \Cref{subsec:results_explainability_embedding}, we aim to go one step further and try to answer the more specific question if there are specialized heads inside of the model architecture for specific prediction tasks, using our IG variant that calculates relevance on the amino acid level per transformer head and layer (see \Cref{sec:algorithm}).

\begin{figure}
\centering
\small
\begin{tabular}{ccc}
\scriptsize{Corr(relev., annot.) $>0$} & \scriptsize{Relevance $>0$} & \scriptsize{Overlay} \\
\includegraphics[width=0.30\columnwidth]{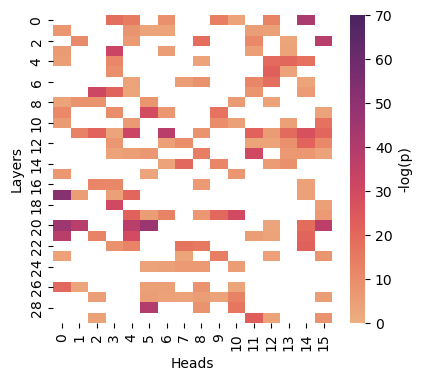} &
\includegraphics[width=0.30\columnwidth]{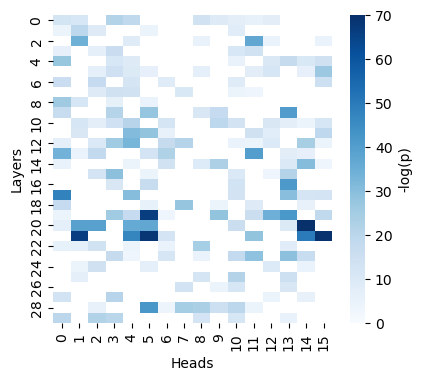} &
\includegraphics[width=0.30\columnwidth]{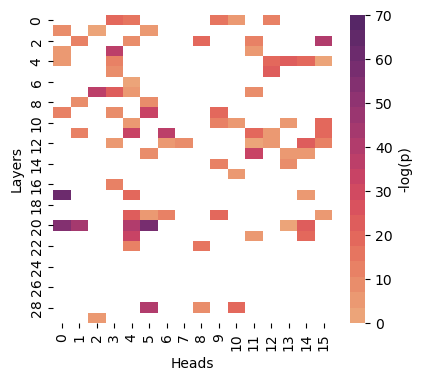} \\
\end{tabular}
\caption{Inside ProtBert; GO ``membrane'' (GO:0016020). 
Left: The relevance attribution (along the sequences) indicative for the GO term ``membrane'' was correlated with UniProt annotations as ``transmembrane regions'', for each transformer head and layer.
Biserial correlation coefficients (r), obtained for each attribution-annotation-pair, were aggregated in population statistics with Wilcoxon signed-rank tests.
The resulting p-values of the tests were adjusted with the Benjamini/Hochberg method for the multiple hypothesis tests conducted in order to limit the false discovery rate. A significance threshold was applied (family-wise error rate of 0.05). 
The negative logarithm of the corrected and thresholded p-values is displayed. All colored pixels indicate statistically significant results.
Center: ProtBert heads with a sig. positive relevance (sum along the sequence; indicative for the GO term ``membrane'') were singled out with the Wilcoxon signed-rank test. The matrix plots show the negative logarithm of the resulting p-values (adjusted with Benjamini/Hochberg and a threshold).
Right: ProtBert heads with a sig. positive attribution-annotation-correlation (p-values from Wilcoxon signed-rank tests plotted) that are also characterized by a sig. positive relevance (the latter overlaid as mask).
Only results for UniProt ``transmembrane regions'' are shown, omitting the results for ``active/binding sites'', ``motifs'', and PROSITE patterns, which did not feature heads with both a sig. positive relevance and attribution-annotation-correlation.}
\label{fig:go_inside_membrane}
\end{figure}

\textbf{GO-prediction: membrane}.
\Cref{fig:go_inside_membrane} shows the results of the explainability analysis inspecting the latent representations inside of the ProtBert model focusing on the selected class of the GO term ``membrane'' (GO:0016020). 
Relevance attributions indicative for GO ``membrane'' per amino acid were correlated with the UniProt annotations as ``transmembrane regions'' separately for each transformer head and layer (matrix plot pixels in \Cref{fig:go_inside_membrane}).
In parallel, ProtBert heads were singled out with a significantly positive relevance (sum along the sequence) indicative for ``membrane'' (see also \Cref{subsec:revealing} and \Cref{sec:algorithm}). 
Both parallel analysis streams were combined by identifying ProtBert heads with both a significantly positive attribution-annotation-correlation and relevance.
Several ProtBert heads in different layers feature a significantly positive correlation of relevance attributions per amino acid with the UniProt annotations as ``transmembrane regions'', going along with a significantly positive relevance for the GO class ``membrane'.
In contrast, correlation of relevance attributions with UniProt ``active'' or ``binding sites'' or ``motifs'' or PROSITE patterns accompanied by a positive relevance was not observed (hence these cases were not included in \Cref{fig:go_inside_membrane}).

\textbf{GO prediction: catalytic activity}.
Supplementary \Cref{fig:go_inside_catalytic} (in Supplementary \Cref{appendix:results_and_discussion}) shows the results of the explainability analysis for the case where the GO term ``catalytic activity'' was selected (GO:0003824). 
Different ProtBert heads stand out characterized by a positive relevance accompanied by a positive correlation of attributions with PROSITE patterns and with UniProt ``active sites'' and ``transmembrane regions'' (but neither with ``binding sites'' nor ``motifs'').

\textbf{GO-prediction: binding}. Supplementary \Cref{fig:go_inside_binding} (in Supplementary \Cref{appendix:results_and_discussion}) repeats the explainability analysis inside ProtBert for the GO term ``binding'' (GO:0005488). For several transformer heads and layers, a positive relevance went along with a correlation of relevance attributions with corresponding PROSITE patterns, and with UniProt ``transmembrane regions'' (but neither with UniProt ``active'' nor ``binding sites'' nor ``motifs'').

\textbf{EC-prediction}. Subsequently, we conducted the explainability analysis for the case where ProtBert had been finetuned to EC number classification on EC50 level L1. Here, the model had learned to differentiate between the six main enzyme classes. Supplementary \Cref{fig:ec_inside} (in Supplementary \Cref{appendix:results_and_discussion}) identifies ProtBert heads characterized both by a positive relevance (sum along the sequence) with respect to the EC class, and by a positive attribution-annotation-correlation (on the amino acid level). The analysis was conducted separately for UniProt annotations as ``active''/``binding sites'', ``transmembrane regions'', and ``motifs'.
(The absence of identified heads for EC4, EC5, and EC6 in the ``transmembrane regions'' rows and for EC1 and EC5 in the ``motif'' rows of Supplementary \Cref{fig:ec_inside} goes along with the availability of relatively few ``transmembrane'' and ``motif'' annotations for these EC classes; see histogram in \Cref{fig:ec_embedding}.)

\textbf{Discussion}.
In summary, we propose a constructive method suited to identify heads inside of the transformer architecture that are specialized for specific protein function or property prediction tasks. 
The proposed method comprises a novel adaptation of the explainable artificial intelligence (XAI) method of IG combined with a subsequent statistical analysis.
We first attributed relevance to the single amino acids per protein (per GO term or EC class), separately for each transformer head and layer. 
Then, we inspected the correlation between relevance attributions and annotations, in a statistical analysis across the annotated proteins from the test split of the respective data set.
Apparently, different transformer heads are sensitive to different annotated and thus biologically and, respectively, chemically ``meaningful'' sites, regions or patterns on the amino acid sequence.

We discuss the benefits of finetuning a pretrained model from end-to-end, and evaluate the XAI method with a residue substitution experiment in Supplementary \Cref{appendix:results_and_discussion}. There, we also discuss the relation of XAI to homology, to the hydrophobicity and charge of residues in transmembrane regions, and to probing and in-silico mutagenesis.

\subsection{Uncovering collective dynamics}

\begin{figure}
\centering
\includegraphics[width=0.5\columnwidth]{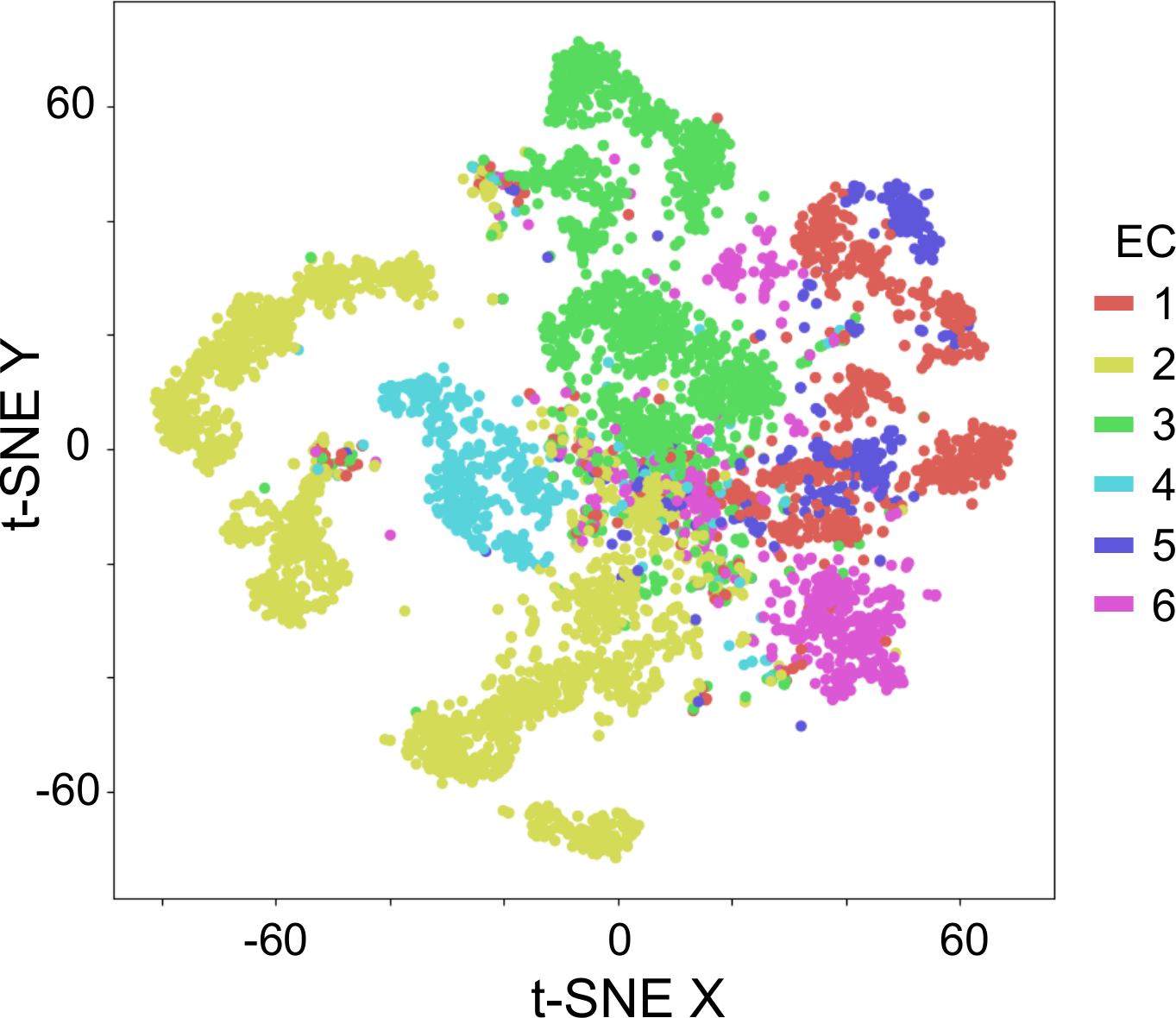}
\caption{PCA and t-SNE visualization of summed attribution maps (ProtBert, EC50, L1).}
\label{fig:ec_tsne}
\end{figure}

Finally, we studied collective dynamics potentially emerging among the transformer heads (ProtBert, EC50, level L1) by a visualization of the originally high-dimensional, summed attribution maps in two dimensions, taking their similarities into account. For this purpose, the attribution maps that were summed along the amino acid sequence and represented as $n_\text{layer}\times n_\text{head}$ matrices (see \Cref{sec:algorithm}) were flattened, resulting in one vector per protein. The dimensionality of these vectors was then reduced with principal component analysis to 50 dimensions, and subsequently to two dimensions with t-distributed stochastic neighbor embedding \citep[t-SNE;][]{vandermaaten2008}, using the default t-SNE parameters. The resulting 2D points were visualized as scatter plot and colored according to the corresponding six main enzyme classes (\Cref{fig:ec_tsne}).

The points form distinctive clusters matching the EC labels. Apparently, a structure emerges in the attribution maps, that seems to indicate class-specific collective dynamics among several ProtBert heads. It is important to stress that the attribution map underlying the clustering no longer contains any reference to specific positions in the sequence but relies on the relevance distribution on the different heads through all layers of the model. The emergence of class-specific structures therefore indicates that there are specific combinations of heads that are relevant for a specific classification decision.

\section{Conclusion}

This work provides additional evidence for the effectiveness of the currently predominant paradigm in deep-learning-based protein analysis through the finetuning of large protein language models from end-to-end (which brings additional benefits; see Supplementary \Cref{appendix:results_and_discussion}). 
For different protein function prediction tasks, this approach leads to best-performing models according to single-model performance. 
The performance level is in many cases on par with MSA-approaches. 
The proposed models can even be effectively combined with the latter through the formation of ensembles.

Considering the ever increasing model complexity, XAI has started to gain traction in the field of protein analysis too \citep{upmeier2019leveraging, taujale2021mapping, vig2021bertology, hou2023learning, vu2023linguistically, zhou2023phosformer}, but quantitative evidence for its applicability beyond single examples was lacking up to now. 
We provide statistical evidence for the alignment of attribution maps with corresponding sequence annotations, both on the embedding level as well as for specific heads inside of the model architecture, which led to the identification of specialized heads for specific protein function prediction tasks.
Emerging class-specific structures suggest that these specialized transformer heads act jointly to decide together in specific combinations.
A further detailed analysis of the identified heads could be an interesting next step in future research, potentially based on the query/key/value (QKV) matrices. Internally to the multi-layered model, a direct correspondence between rows/columns of the QKV matrices and individual residues in the sequence is, however, not possible anymore. This limitation makes it, e.g. difficult to infer relations between residues from the QKV matrices.

In summary, XAI promises to tap into the presumably substantial knowledge contained in large models pretrained on massive data sets of amino and/or nucleic acid sequences \citep{steinegger2019protein, ji2021dnabert}.
Therefore, we expect that XAI will play an increasingly important role in the future of bioinformatics research.
We see potential applications of XAI for model validation and for scientific discovery (e.g. of novel discriminative sequence patterns or motifs that have not been identified by experiments or MSA so far). 
Identifying specialized heads might also help to prune overly large models, making them smaller and more efficient.

\vspace*{-10pt}
\section*{Conflict of interest}
None declared.

\section*{Funding}
This work was supported by the Bundesministerium f\"ur Bildung und Forschung through the BIFOLD - Berlin Institute for the Foundations of Learning and Data [grant numbers 01IS18025A, 01IS18037A].

\vspace*{-12pt}

\bibliographystyle{natbib}
\small{\bibliography{bibfile}}
\normalsize
\clearpage
\pagenumbering{roman}
\appendix
\renewcommand{\thesection}{\Alph{section}}
\renewcommand{\thesubsection}{\Roman{subsection}}
\counterwithin{figure}{section}
\counterwithin{table}{section}

\section{Supplementary material: Introduction} 
\label{appendix:introduction}

\begin{table}[!ht]
\caption{
Definitions of UniProt annotations and of corresponding GO terms quoted from the \href{https://www.uniprot.org/}{UniProt} and \href{https://www.ebi.ac.uk/QuickGO/}{EMBL-EBI} websites.}
\label{tab:uniprot_go}
\small
\begin{tabular}{p{1.5in} p{1.4in}}
\toprule
\Centering{UniProt annotation} & \Centering{GO term} \\
\Centering{(per amino acid)} & \Centering{(per protein)} \\
\midrule
\RaggedRight{\href{https://www.uniprot.org/help/act_site}{\textbf{Active site}}: ``used for enzymes and indicates the residues directly involved in catalysis''} &
\RaggedRight{\href{https://www.ebi.ac.uk/QuickGO/term/GO:0003824}{\textbf{Catalytic activity} (GO:0003824)}: ``Catalysis of a biochemical reaction [...] catalysts are naturally occurring macromolecular substances known as enzymes [that] possess specific binding sites for substrates.''} \\
\midrule
\RaggedRight{\href{https://www.uniprot.org/help/binding}{\textbf{Binding site}}: ``interaction between protein residues and a chemical entity''} &
\RaggedRight{\href{https://www.ebi.ac.uk/QuickGO/term/GO:0005488}{\textbf{Binding} (GO:0005488)}: ``selective, non-covalent, often stoichiometric, interaction of a molecule with one or more specific sites on another molecule.''} \\
\midrule
\RaggedRight{\href{https://www.uniprot.org/help/transmem}{\textbf{Transmembrane region}}: ``extent of a membrane-spanning region of the protein. [...] both alpha-helical transmembrane regions and the membrane spanning regions of beta-barrel transmembrane proteins.''} &
\RaggedRight{\href{https://www.ebi.ac.uk/QuickGO/term/GO:0016020}{\textbf{Membrane} (GO:0016020)}: ``A lipid bilayer along with all the proteins and protein complexes embedded in it an attached to it.''} \\
\midrule
\RaggedRight{\href{https://www.uniprot.org/help/motif}{\textbf{Short sequence motif}}: ``a short (usually not more than 20 amino acids) conserved sequence motif of biological significance. Specific sequence motifs usually mediate a common function, such as protein-binding or targeting to a particular subcellular location, in a variety of proteins.''} &
\RaggedRight{--} \\
\bottomrule
\end{tabular}
\end{table}

\section{Supplementary material: System and methods} 
\label{appendix:system_and_methods}

\subsection{Protein function prediction tasks and data sets}
\label{subsec:tasks}

\textbf{GO prediction: introduction}. We first focus on the protein function prediction task of inferring Gene Ontology (GO) terms, that the proteins are labeled with, from their amino acid sequence. This particularly comprehensive task allows characterizing each protein with respect to several properties, taking its molecular function, location in the cell, and involvement in different biological processes into account. GO term prediction has been approached using a range of deep learning methods in the past years \citep[e.g. ][]{kulmanov2017deepgo, kulmanov2019deepgoplus, kulmanov2022deepgozero, You2018Golabeler, you2018deeptext2go, you2021deepgraphgo, strodthoff2020udsmprot, littmann2021embeddings}. 
The \href{http://geneontology.org/}{GO consortium} \citep{ashburner2000gene,GO2020} aims at representing biological knowledge in the form of an ontology, i.e. as a graph comprising terms/classes as nodes and the relationships between them as edges. Actually, GO consists of three separate ontologies for molecular function (MFO), cellular component (CCO) and biological process (BPO). The protein database UniProtKB/Swiss-Prot \citep{UniProt2021} lists the amino acid sequences of proteins, and labels these proteins with selected, corresponding GO terms. In addition, the database contains selected annotations on the amino acid level, such as amino acids that serve as active or binding sites (see \Cref{tab:uniprot_go}), which will later play a role in this work.

\textbf{GO prediction: setup}. Considering that each protein can typically be labeled with several of the numerous GO terms, the task was phrased here as a multi-label (``protein-centric'') prediction problem, where the model was trained to produce a label vector of about 5000 dimensions (5220 or 5101; see below) for each protein. The mapping from GO terms to the intricate topology of the ontology had been simplified by \citet{kulmanov2019deepgoplus}, who made the data available in their accompanying \href{http://deepgoplus.bio2vec.net/data/}{data repository}, as follows.
A ``flat'' label vector represents the original graph structure of the ontology. Labels were propagated from children terms to parents through the hierarchy of the ontology (in contrast to the original GO, where labels are assigned to ontology leaves as far as possible). Only a subset of about 5k of the about 45k GO terms in total was considered, because the GO terms had been filtered to those that occur at least 50 times in the respective data set. Like in \citet{kulmanov2019deepgoplus}, predictions are propagated towards the root node of the respective ontology tree. Evaluation is always performed with respect to the set of all GO terms, where the model can only predict the most frequently occurring GO terms.

\textbf{GO prediction: data sets}. We used both the ``CAFA3'' and the ``2016'' data sets by \citet{kulmanov2019deepgoplus} with 66841 training and 3328 testing samples with 5220 classes (``CAFA3'') and, respectively, with 65028 training and 1788 testing samples with 5101 classes (``2016'') \citep[see Section~2.1 with Table~1 in][]{kulmanov2019deepgoplus}. While the ``CAFA3'' data set is a widely used benchmark \citep{zhou2019cafa}, the time-based ``2016'' splits of the underlying UniProtKB/Swiss-Prot database serve for the comparison with the results by \citet{You2018Golabeler, you2018deeptext2go, kulmanov2019deepgoplus, strodthoff2020udsmprot} (``2016'' because entries from Jan.-Oct. 2016 serve for testing). The predictive performance of the protein function prediction models was assessed separately for the molecular function, cellular component and biological process ontologies of GO, with the $F_\text{max}$ and $S_\text{min}$ metrics \citep{clark2013information} used in ``CAFA3'' \citep{zhou2019cafa}, and with the area under the precision-recall (AUPR) curve \citep[comparable to ][]{you2018deeptext2go, kulmanov2019deepgoplus, You2018Golabeler, strodthoff2020udsmprot}.

\textbf{EC prediction}. During the course of the present study, we narrow down the prediction task to the inference of the enzymatic function of a protein from its amino acid sequence.
Enzymes are proteins that catalyze chemical reactions, which can be grouped into seven main classes and further sub-classes, as categorized by the Enzyme Commission (EC) numbers, a nomenclature developed by the International Union of Biochemistry and Molecular Biology \citep{webb1992enzyme, mcdonald2008explorenz}. 
EC prediction has been approached with machine learning, e.g. by \citet{dalkiran2018ecpred, li2018deepre, zou2019mldeepre, strodthoff2020udsmprot, yu2023enzyme}.

The studied prediction task comprised the binary classification of whether a protein serves as enzyme or not (EC level ``L0''), the classification into the main classes of enzymatic reactions (EC level ``L1''), and the differentiation among different sub-classes of enzymatic reactions (EC level ``L2''). 
We assessed the predictive performance on the ``EC40'' (58018 samples) and ``EC50'' (104940 samples) data sets by \citet{strodthoff2020udsmprot} (see their Supplementary Section S1), using accuracy as metric. The numbers in ``EC40'' and ``EC50'' refer to the similarity threshold between train and test splits of 40\% similarity (more difficult task) and, respectively, 50\% (easier task).
The data set comprises the traditional six main EC classes (oxidoreductases, transferases, hydrolases, lyases, isomerases, ligases) leaving apart the new seventh EC class of translocases.

\textbf{Data availability and composition of the training set}. 
GO data by \citet{kulmanov2019deepgoplus} from their \href{https://deepgo.cbrc.kaust.edu.sa/data/}{data repository}
and EC data by \citet{strodthoff2020udsmprot} (see their Supplementary Section S1) were preprocessed as detailed in the \href{https://github.com/nstrodt/UDSMProt}{code repository} by \citet{strodthoff2020udsmprot}, resulting in separate data sets for GO ``2016'' (a.k.a. ``temporalsplit''), for GO ``CAFA3'', and for EC40 and EC50 on levels L0, L1, and L2.
Dedicated validation (a.k.a. development) splits were available for the EC and GO ``CAFA3'' tasks. For the time-based GO ``2016'' split, a 10~\% subset was randomly sampled from the training data and used as validation split \citep[comparable to ][]{kulmanov2019deepgoplus}. As described below, we use the validation set for model selection and report the test set score.

\subsection{Finetuning pretrained transformers}
\label{subsec:finetuning}

\textbf{Considered transformer models}. We finetuned the transformer model ``ProtBert-BFD'' \citep{Elnaggar2022} to the GO term and EC number prediction tasks detailed earlier in Section \ref{subsec:tasks} of \Cref{appendix:system_and_methods}. ``ProtBert-BFD'' is a variant of BERT \citep{devlin2018bert}, with 420 million parameters distributed among 16 heads in 30 layers, that had been pretrained in a self-supervised way with a masked token prediction task on the massive metagenomic protein sequence collection titled ``Big Fantastic Database'' (BFD), which comprises more than two billion sequences \citep{jumper2021highly, steinegger2019protein, steinegger2018clustering}.

In addition, we finetuned also the more complex and memory intensive ``ProtT5-XL-UniRef50'' model \citep{Elnaggar2022} on the GO and EC tasks, using the last hidden state of the encoder.
`ProtT5-XL-UniRef50'' is based on the ``Text-To-Text Transfer Transformer'' \citep[T5; ][]{colin2020t5} with about three billion parameters (1.2 billion for the encoder used here) and had been pretrained on 45 million sequences from UniRef50 \citep{suzek2014uniref}.
We abbreviate ``ProtBert-BFD'' as ``ProtBert'' and ``ProtT5-XL-UniRef50'' as ``ProtT5'' in the following text.

Moreover, we finetuned two ESM-2 \citep{lin2023evolutionary} variants, both pretrained on UniRef50. The relatively small ESM-2 variant ``esm2\_t6\_8M\_UR50'' comprises 6 layers, 8M parameters and an embedding dimension of 320. The larger variant ``esm2\_t33\_650M\_UR50D'' (33 layers, 650M parameters, embedding dimension of 1280) was chosen, because its size is roughly comparable to ProtBert-BFD (30 layers, 420M parameters, 1024 encoder features) and ProtT5-XL-UniRef50 (24 layers, 1200M parameters, 1024 encoder features).

\textbf{Classification head}. The final classification head comprised a hidden layer and then a linear layer as connection to the dimensionality of the respective label vector. A hidden layer with $n=50$ neurons was chosen for the simpler EC task. For the more demanding GO task, the size of the hidden layer was increased to $n=2\times 5220$ (GO ``CAFA3'') and, respectively, to $n=2\times 5101$ (GO ``2016''), such that the hidden layer contained twice as many neurons in comparison to the dimensionality of the respective label vector. Rectified linear units served as activation function, and layer normalization and a dropout layer with a dropout rate of ten percent served for regularization.

\textbf{Finetuning procedure}. For finetuning, the input sequence was cropped to a maximum sequence length of 1000 amino acids per protein, if necessary.
The encoder features created by the transformer were pooled by concatenating the classification token, the maximum and the mean of the features along the sequence, and the sum of the features divided by the square root of the sequence length \citep{reimers2019sentencebert}.
Binary cross entropy combined with a sigmoid layer served as loss function for the multi-label GO task, and cross entropy loss combined with softmax for the multi-class EC task. Gradients were accumulated over 64 batches of size one each.
Adam \citep{kingma2014adam} was selected as optimizer. The learning rate was set to $5\times10^{-6}$ for the encoder, which was frozen for the first epoch, and to $3\times10^{-5}$ for the classification head. For the multi-label GO task, the rate was additionally scheduled with a linear increase over 500 warm-up steps in the beginning and then a decrease to zero following a half-cosine function calculated for a total number of 20000 training steps.

\textbf{Model selection}. After each training epoch, the loss (GO) or accuracy (EC) on the validation split was monitored and the model with the minimum loss (GO) or maximum accuracy (EC) was saved. Training was repeated until no further improvement was observed on the validation split. The best model was typically found after around 2-13 epochs in the case of ProtBert, after around 2-7 epochs in the case of ProtT5, after 6-20 epochs for the smaller, and 2-7 epochs for the larger ESM-2 variant.

\section{Supplementary material: Algorithm} 
\label{appendix:algorithm_summed_attribution_maps}

\textbf{Positive attributions.} For the summed attribution maps (but not for the correlation analysis), the positive attributions are of higher importance, since this analysis focuses on identifying the areas of a sequence that speak for a specific class, rather than finding the areas that speak against this sequence being classified as such. Most XAI cases, such as the multiple classification attribution examples in \citet{sundararajan2017}, choose this approach. Some granularity is being lost because the attributions of individual amino acids are aggregated. The case that a head identifies the relevant areas with a positive attribution, and those values are canceled out by also finding negative-relevant areas is theoretically possible, even though a positive relevance sum is an indicator for the respective class, because the relevance sign is not arbitrary \citep{binder2023shortcomings}. Nevertheless, we account for a potentially too rigid filtering by showing the correlation analysis results and the summed attribution maps individually.

\section{Supplementary material: Implementation} 
\label{appendix:implementation}

The pretrained models of ProtBert, ProtT5 \citep{Elnaggar2022}, and ESM-2 \citep{lin2023evolutionary} were connected to an additional pooling layer and classification head (see \Cref{appendix:system_and_methods}, \ref{subsec:finetuning}), and finetuned on GO data provided by \citep{kulmanov2019deepgoplus} and, respectively, EC data by \citet{strodthoff2020udsmprot} (see their Supplementary Section S1), using PyTorch \citep{Paszke2019} with \href{https://github.com/Lightning-AI/lightning}{PyTorch Lightning} as wrapper. 
The adaptation of the integrated gradients algorithm acting separately on each transformer head in every layer (see \Cref{sec:algorithm}) was implemented in Python building on the layer integrated gradients method of Captum \citep{captum}.
Annotations were obtained from the databases UniProt \citep{UniProt2021} and PROSITE \citep{sigrist2012new} in combination with GO \citep{ashburner2000gene,GO2020}. 
Statistical calculations were performed using SciPy \citep{scipy} and statsmodels \citep{seabold2010statsmodels}.
Performance on the GO task was assessed based on an adaptation by \citep{strodthoff2020udsmprot} from \citep{kulmanov2019deepgoplus}. 
Source code for model training, model interpretation, statistics and visualization can be accessed at \url{https://github.com/markuswenzel/xai-proteins}.

\section{Supplementary material: Results and Discussion} 
\label{appendix:results_and_discussion}

\textbf{Research question}. Here, we aim to answer the question whether finetuning of large pretrained transformer models leads to a competitive performance in protein function prediction.

\textbf{GO prediction on the CAFA3 data set}. \Cref{tab:go_cafa3} shows the performance results for GO term prediction by finetuning and testing ProtBert, ProtT5 and ESM-2 on the ``CAFA3'' splits. Performance was assessed with the protein-centric $F_\text{max}$ metric, separately for the molecular function (MFO), biological process (BPO), and cellular component ontologies (CCO) that comprise the GO. For comparison with the state-of-the-art, we added results reported in the literature \citep{olenyi2023, bernhofer2021, littmann2021embeddings, kulmanov2019deepgoplus, You2018Golabeler} to \Cref{tab:go_cafa3}. 
For model evaluation, we distinguish between single-models and ensemble models. 
In the former category, the finetuned ProtT5 turns out to be the model that outperforms both prior deep-learning-based approaches based on transformers (goPredSim) and convolutional neural networks (DeepGOCNN) as well as the very strong MSA-based baseline DiamondScore, on all ontologies MFO, BPO and CCO. The finetuned larger ESM-2 variant is a comparably strong competitor to ProtT5 (and better than ProtT5 on BPO).

Turning to ensemble model, we find that the ensemble of a finetuned ProtBert (followed by ProtT5) and DiamondScore \citep[MSA-based; ][]{kulmanov2019deepgoplus} produces the best predictions on CCO \citep[scores were combined with a weighted sum; comparable to ][see their Section~2.3]{kulmanov2019deepgoplus}. 
The ensemble of the finetuned larger ESM-2 variant with DiamondScore leads the ranking on BPO.
The ensemble method ``GOLabeler'' \citep{You2018Golabeler} leads on MFO. GOLabeler had won the original CAFA3 challenge on all three ontologies BPO, CCO, and MFO \citep[see Fig.~3 of ][where the $F_\text{max}$ results are reported]{zhou2019cafa}.

\begin{table*}
\caption{
GO term prediction by finetuning and testing ProtT5, ProtBert, and ESM-2 on the ``CAFA3'' splits, next to state-of-the-art results reported in the literature. Predictive performance was evaluated with the protein-centric $F_\text{max}$ metric. The molecular function, biological process and cellular component ontologies of GO are abbreviated as MFO, BPO and CCO. Best single-model results are marked in bold face. Best overall results are underlined. Arrows ($\rightarrow$) mark results achieved in this work. ProtT5 provides the best overall single-model performance in the MFO and CCO categories, while the larger ESM-2 variant performs best on BPO. Ensembling with the MSA-based DiamondScore predictions is highly effective.}
\label{tab:go_cafa3}
\small \centering
\begin{tabular}{clccc}
\toprule
& Method & MFO, $F_\text{max}$ (($\uparrow$) & BPO, $F_\text{max}$ ($\uparrow$)& CCO, $F_\text{max}$ ($\uparrow$)\\ 
\midrule
\multirow{8}{*}{\rotatebox[origin=c]{90}{Single models}}
& $\rightarrow$ ProtT5-XL-UniRef50 \citep{Elnaggar2022}, finetuned & \textbf{0.523} & 0.442 & \underline{\textbf{0.641}} \\
& $\rightarrow$ ProtBert-BFD \citep{Elnaggar2022}, finetuned & 0.503 & 0.441 & 0.630 \\
& $\rightarrow$ esm2\_t33\_650M\_UR50D \citep{lin2023evolutionary}, finetuned & 0.512 & \textbf{0.447} & 0.639 \\
& $\rightarrow$ esm2\_t6\_8M\_UR50D \citep{lin2023evolutionary}, finetuned & 0.503 & 0.419 & 0.629 \\
& goPredSim, ProtT5 embeddings \citep{olenyi2023} & \textbf{0.52}$\pm$3\% & 0.38$\pm$2\% & 0.59$\pm$2\% \\
& goPredSim, SeqVec embeddings \citep{bernhofer2021} & \textbf{0.52}$\pm$2\% & 0.37$\pm$2\% & 0.58$\pm$2\% \\
& goPredSim, SeqVec embeddings \citep{littmann2021embeddings} & 0.50$\pm$3\% & 0.37$\pm$2\% & 0.57$\pm$2\% \\
& DeepGOCNN \citep{kulmanov2019deepgoplus} & 0.420 & 0.378 & 0.607 \\
& DiamondScore \citep{kulmanov2019deepgoplus} & 0.509 & 0.427 & 0.557 \\ 
\midrule
\multirow{8}{*}{\rotatebox[origin=c]{90}{Ensembles}}
& $\rightarrow$ ProtT5-XL-UniRef50 \citep{Elnaggar2022}, finetuned + DiamondScore & 0.565 & 0.473 & 0.635 \\
& $\rightarrow$ ProtBert-BFD \citep{Elnaggar2022}, finetuned + DiamondScore & 0.580 & \underline{0.480} & 0.640 \\
& $\rightarrow$ esm2\_t33\_650M\_UR50D \citep{lin2023evolutionary}, finetuned + DiamondScore & 0.566 & \underline{0.480} & 0.634 \\
& $\rightarrow$ esm2\_t6\_8M\_UR50D \citep{lin2023evolutionary}, finetuned + DiamondScore & 0.559 & 0.470 & 0.629 \\
& DeepGOPlus \citep{kulmanov2019deepgoplus} & 0.544 & 0.469 & 0.623 \\
& GOLabeler \citep{You2018Golabeler}, \citep[Fig.~3 of ][]{zhou2019cafa} & \underline{0.62} & 0.40 & 0.61 \\
\bottomrule
\end{tabular}
\end{table*}

\textbf{GO prediction on the 2016 data set}.
\Cref{tab:go_2016} shows the GO term prediction results of the finetuned ProtT5, ProtBert, and ESM-2 models on the more recent ``2016'' data set \citet{kulmanov2019deepgoplus, You2018Golabeler}, as well as the literature results \citep[compiled by ][ which report AUPR and $S_\text{min}$ results in addition to $F_\text{max}$]{kulmanov2019deepgoplus, strodthoff2020udsmprot}.
Again, we distinguish between single and ensemble models in the following discussion.

Concerning single-models, different methods obtain the best results, depending on the considered metric and ontology.
ProtT5 is the best single-model as measured by AUPR on all three ontologies MFO, BPO, CCO. 
ProtT5 also obtained the largest $F_\text{max}$ on BPO while the larger ESM-2 variant is best on CCO.
DiamondScore, an MSA-based method, is the best single-model with respect to the $S_\text{min}$ metrics on the ontologies MFO and BPO, and achieves the best $F_\text{max}$ on MFO. 
DeepGOPlus has the lowest and thus best $S_\text{min}$ for CCO.

Turning to ensemble models, we find that the best overall results are obtained again by different ensemble methods. 
The finetuned ProtT5 combined with DiamondScore leads with respect to the AUPR metric and to $F_\text{max}$ on BPO.
The larger ESM-2 combined with DiamondScore leads the AUPR-ranking on CCO.
Both ProtT5 and DeepGOPlus have the best $F_\text{max}$ on CCO. 
DeepText2GO is best on MFO according to $F_\text{max}$ and AUPR, and on CCO according to $S_\text{min}$. 
GOLabeler features the smallest and thus best $S_\text{min}$ on MFO and BPO.

\begin{table*}
\caption{
GO term prediction by finetuning and testing ProtT5, ProtBert, and ESM-2 on the ``2016'' data set splits \citet{kulmanov2019deepgoplus,You2018Golabeler}, next to the literature results collected by \citet{kulmanov2019deepgoplus} and \citet{strodthoff2020udsmprot}. Best single-model results are marked in bold face (highest $F_\text{max}$ and AUPR; lowest $S_\text{min}$). Best overall results are underlined. Arrows ($\rightarrow$) point towards models finetuned in this paper, with competitive results.}
\label{tab:go_2016}
\footnotesize
\begin{tabular}{ cl ccc | ccc | ccc}
\toprule
& & \multicolumn{3}{c|}{$F_\text{max}$ ($\uparrow$)} & \multicolumn{3}{c|}{$S_\text{min}$ ($\downarrow$)} & \multicolumn{3}{c}{AUPR ($\uparrow$)}\\	
& Method & MFO & BPO & CCO & MFO & BPO & CCO & MFO & BPO & CCO\\
\midrule
\multirow{6}{*}{\rotatebox[origin=c]{90}{Single-models}}
& $\rightarrow$ ProtT5-XL-UniRef50 \citep{Elnaggar2022}, finetuned & 0.526 & \textbf{0.441} & 0.685 & 10.099 & 35.229 & 8.030 & \textbf{0.487} & \textbf{0.395} & \textbf{0.730}\\
& $\rightarrow$ ProtBert-BFD \citep{Elnaggar2022}, finetuned & 0.503 & 0.424 & 0.677 & 10.421 & 35.401 & 8.198 & 0.460 & 0.369 & 0.714 \\
& $\rightarrow$ esm2\_t33\_650M\_UR50D \citep{lin2023evolutionary}, finetuned & 0.507 & 0.433 & \textbf{0.688} & 10.256 & 35.291 & 8.124 & 0.466 & 0.376 & 0.726 \\
& $\rightarrow$ esm2\_t6\_8M\_UR50D \citep{lin2023evolutionary}, finetuned & 0.520 & 0.423 & 0.681 & 10.016 & 35.618 & 8.199 & 0.486 & 0.371 & 0.715 \\
& UDSMProt \citep{strodthoff2020udsmprot} &	0.481 & 0.411 & 0.682 & 10.505 & 36.147 & 8.244 &	0.472& 0.356 & 0.704\\
& DeepGO \citep{kulmanov2017deepgo} & 0.449 &	0.398 &	0.667 &	10.722 & 35.085 & \textbf{7.861} & 0.409 & 0.328 &	0.696 \\
& DeepGOCNN \citep{kulmanov2019deepgoplus} & 0.409 & 0.383 & 0.663 & 11.296 & 36.451 & 8.642 & 0.350 & 0.316 & 0.688\\
& DiamondScore \citep{kulmanov2019deepgoplus} & \textbf{0.548} & 0.439 & 0.621 & \textbf{8.736} & \textbf{34.060} & 7.997 & 0.362 & 0.240 & 0.363 \\
\midrule
\multirow{6}{*}{\rotatebox[origin=c]{90}{Ensembles}}
& $\rightarrow$ ProtT5-XL-UniRef50 \citep{Elnaggar2022}, finetuned + $\Diamond$ & 0.591 & \underline{0.482} & \underline{0.699} & 8.583 & 33.752 & 7.467 & 0.571 & \underline{0.431} & 0.736 \\
& $\rightarrow$ ProtBert-BFD \citep{Elnaggar2022}, finetuned + $\Diamond$ & 0.583 & 0.470 & 0.693 & 8.687 & 33.560 & 7.597 & 0.558 & 0.423 & 0.725 \\
& $\rightarrow$ esm2\_t33\_650M\_UR50D \citep{lin2023evolutionary}, finetuned + $\Diamond$ & 0.594 & 0.476 & 0.697 & 8.656 & 33.498 & 7.583 & 0.564 & 0.427 & \underline{0.737}\\
& $\rightarrow$ esm2\_t6\_8M\_UR50D \citep{lin2023evolutionary}, finetuned + $\Diamond$ & 0.598 & 0.476 & 0.696 & 8.600 & 33.576 & 7.542 & 0.567 & 0.424 & 0.729\\
& UDSMProt \citep{strodthoff2020udsmprot} + $\Diamond$ & 0.582 & 0.475 & 0.697 & 8.787 & 33.615 & 7.618	& 0.548 & 0.422 & 0.728\\
& DeepGOPlus \citep{kulmanov2019deepgoplus} & 0.585 & 0.474 & \underline{0.699} & 8.824 & 33.576 & 7.693 & 0.536 & 0.407 & 0.726\\
& DeepText2GO \citep{you2018deeptext2go} & \underline{0.627} &	0.441 &	0.694 &	5.240 &	17.713 & \underline{4.531} & \underline{0.605} & 0.336 &	0.729\\ 
& GOLabeler \citep{You2018Golabeler} &	0.580 &	0.370 &	0.687 &	\underline{5.077} &	\underline{15.177} & 5.518 &	0.546 &	0.225 &	0.700 \\
\bottomrule
\end{tabular}
{$\Diamond$: DiamondScore}
\end{table*}

\textbf{EC prediction}.
\Cref{tab:ec_exp} shows how accurately ProtBert, ProtT5, and ESM-2 predicted the EC number of a given protein from its amino acid sequence, separately for the three EC levels (L0, L1, L2) and two train/test splits (EC40, EC50). For comparison, the table lists the results of ``UDSMProt'' too \citep{strodthoff2020udsmprot}. 
The transformer models predict the EC number more accurately, with a large gap, in comparison to the RNN-based UDSMProt.
ProtBert, ProtT5, and the larger ESM-2 variant are comparable with respect to their strong performance.

\begin{table*}
\caption{
The EC number of each protein was predicted from the amino acid sequence. Different EC levels  (L0: enzyme or not, L1: six main enzyme classes, L2: enzyme sub-classes) and train/test splits (EC40/EC50) were assessed separately, with accuracy as performance metric. ProtT5, ProtBert and ESM-2 transformers were finetuned and the results compared to the RNN-based ``UDSMProt'' \citep[cf. ][Table~1; incl. baseline of a convolutional neural network (CNN) operating on MSA features]{strodthoff2020udsmprot}. Arrows ($\rightarrow$) mark results established here.}
\label{tab:ec_exp}
\small \centering
\begin{tabular}{l lll|lll}
\toprule
& \multicolumn{3}{c|}{EC40} & \multicolumn{3}{c}{EC50}\\
Model & L0 & L1 & L2 & L0 & L1 & L2\\
\midrule
$\rightarrow$ ProtT5-XL-UniRef50 \citep{Elnaggar2022}, finetuned & \textbf{0.966} & \textbf{0.968} & \textbf{0.954} & 0.985 & \textbf{0.993} & \textbf{0.985}\\
$\rightarrow$ ProtBert-BFD \citep{Elnaggar2022}, finetuned & 0.960 & 0.958 & 0.945 & 0.981 & 0.990 & 0.982\\
$\rightarrow$ esm2\_t33\_650M\_UR50D \citep{lin2023evolutionary}, finetuned & \textbf{0.966} & 0.961 & 0.942 & \textbf{0.986} & 0.992 & 0.983\\
$\rightarrow$ esm2\_t6\_8M\_UR50D \citep{lin2023evolutionary}, finetuned & 0.926 & 0.855 & 0.811 & 0.958 & 0.943 & 0.929\\
UDSMProt (``Fwd+bwd; pretr.'') \citep{strodthoff2020udsmprot} & 0.91 & 0.87 & 0.84 & 0.96 & 0.97 & 0.95\\
CNN (``Baseline, seq'') \citep{strodthoff2020udsmprot} & 0.84 & 0.61 & 0.47 & 0.92 & 0.80 & 0.79 \\
\bottomrule
\end{tabular}
\end{table*}

\textbf{Discussion}.
Finetuning large, pretrained protein language models on the tasks of GO term as well as EC number prediction results in a strong predictive performance that is competitive with, and in several cases, better than state-of-the-art methods from the literature (depending on the metric and ontology under consideration, in the GO case). This outcome highlights again the benefits of transferring pretrained universal language models to downstream tasks in the proteomics field \citep[cf., e.g. ][]{strodthoff2020udsmprot,Rao2019,Elnaggar2022}.

The largest inspected models lead the rankings on different metrics (and ontologies), with the bigger ESM-2 being a strong competitor to ProtT5.
Again, ProtT5 consistently ranks better than (the smaller) ProtBert both on ``CAFA3'' and on the more recent, harder ``2016'' benchmark (only the ProtBert-DiamondScore ensemble forms an exception on ``CAFA3'' with respect to $F_\text{max}$ and on ``2016'' with respect to $S_\text{min}$ in the BPO case).

DeepGOCNN and DiamondScore are competing single-models with results available from the literature on both benchmarks.
ProtT5, ProtBert, and ESM-2 outperform DeepGOCNN in both benchmarks on all metrics. 
ProtT5 and (in particular, the larger variant of) ESM-2 outperform the strong DiamondScore, which is based on MSAs, in both benchmarks in many cases.

The approach of finetuning the entire model including the encoder shows its strength in the ``CAFA3'' benchmark for BPO and CCO, where the finetuned ProtT5 outperforms the ``goPredSim'' model which is based on nearest-neighbor-lookup using features extracted from ProtT5 embeddings \citep{olenyi2023}. The observation that this nearest-neighbor-lookup is competitive in the MFO ontology might be related to the relatively dense annotation of the MFO.

In summary, we showed that finetuning pretrained large transformer models leads to competitive results for protein function prediction tasks, in particular in the most relevant comparison in the single-model category.

\textbf{On pretraining and finetuning from end-to-end.}
First of all, it is important to stress that our approach can only be applied to a model that includes a classification head as it relies on the class-specific output prediction. We studied how finetuning the entire model (comprising the pretrained ProtBert and the classification head) from end-to-end compares to training only the classification head while keeping the pretrained ProtBert frozen i.e. unchanged. We refer to these two scenarios as ``finetuned'' vs. 'pretrained'. We also considered an additional baseline where the ProtBert parameters were ``shuffled'' per layer, hence keeping realistic weight statistics across layers.
Only the classification head was trained in the ``shuffled''  scenario, while the encoder parameters were kept frozen. For comparability, we trained the models in all scenarios for the same number of epochs, i.e. until early stopping was initiated in the ``finetuned'' scenario.

First, we inspected the performance results for these scenarios. 
For the GO term classification task ``2016'', the finetuned ProtBert (see \Cref{tab:go_2016}) achieved an $F_\text{max}$ of 0.503 on MFO, of 0.424 on BPO, and of 0.677 on CCO. Performance dropped when only the classification head was trained, while the pretrained encoder was kept frozen, to an $F_\text{max}$ of 0.463 on MFO, of 0.407 on BPO, and of 0.674 on CCO. Parameter shuffling resulted in an $F_\text{max}$ of 0.305 on MFO, of 0.316 on BPO, and of 0.601 on CCO (close to the ``naive'' baseline from Table~3 of \citet{strodthoff2020udsmprot}).

For the enzyme classification task ``EC50 level L1'', the accuracy of 0.990 of the finetuned ProtBert, see \Cref{tab:ec_exp}, dropped to an accuracy of 0.829 achieved by training only the classification head on top of the pretrained but frozen ProtBert. Performance dropped further to 0.369 in the ``shuffled'' scenario.

Thus, finetuning from end-to-end apparently provides advantages in comparison to only training the classification head on top of the pretrained encoder. Using pretrained parameters in both the ``finetuned'' and the ``pretrained'' scenarios shows the well-known benefits of transfer learning in comparison to the ``shuffled'' scenario.

\begin{figure}
\centering
\begin{tabular}{cccc}
\includegraphics[trim={0 0 2cm 0}, clip, height=0.3\columnwidth]{membrane-overlay-pos-rel-with-annot-rel-corr-pbr-transmembrane.png} &
\includegraphics[trim={0 0 2cm 0}, clip, height=0.3\columnwidth]{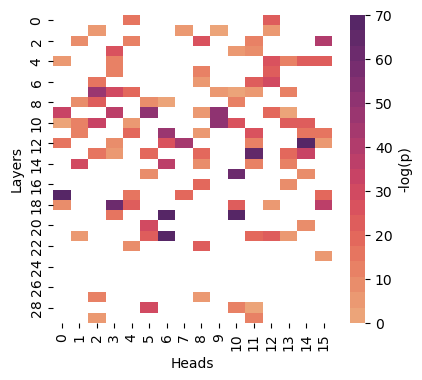} & 
\includegraphics[height=0.3\columnwidth]{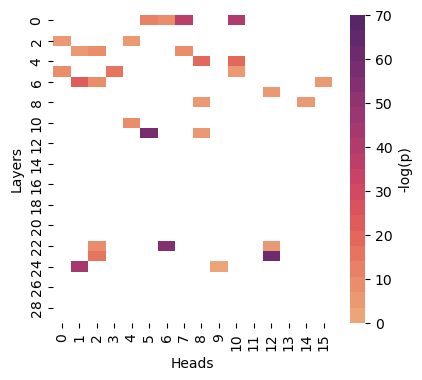} \\
\end{tabular}
\caption{Attribution analysis results were similar in the comparable ``finetuned'' (left) and ``pretrained'' (centre) scenarios, but different in the disparate ``shuffled'' (right) scenario, as expected (inside ProtBert, GO membrane/transmembrane). Jaccard similarity of the overlap of ``significant heads'' between scenarios was 0.43 for the finetuned-pretrained, 0.036 for the finetuned-shuffled, and 0.035 for the pretrained-shuffled pair.}
\label{fig:finetuned_pretrained_shuffled}
\end{figure}

The corresponding attribution analysis results (see \Cref{fig:finetuned_pretrained_shuffled}) for the ``finetuned'' and the ``pretrained'' scenario have much in common -- in contrast to the ``shuffled'' scenario, which shows a disparate picture. Overlap of ``significant heads'' between each scenario pair was quantified with the Jaccard similarity coefficient as 0.43 for the finetuned-pretrained pair, in contrast to the smaller 0.036 for the finetuned-shuffled pair, and 0.035 for the pretrained-shuffled pair. Both ``finetuned'' and ``pretrained'' scenarios build upon the same pretrained foundation model. Therefore, a similar attribution analysis result can be expected. When the parameters were ``shuffled'', fewer heads appear in the plot. This outcome shows that the XAI method is sensitive to the model parameters, which is expected and required \citep{adebayo2018sanity}. It varies across different training runs, whereas the finetuned results remain largely similar (as already suggested by the large similarity of full finetuning vs. pretrained). It is noteworthy to stress that on general grounds the plot may still show significant heads due to a coincidental partial alignment of certain randomly initialized heads with directions in feature space that happen to correlate spatially with sequence annotations. Besides, skip connections are not affected by the randomization \citep{binder2023shortcomings}. Model parameter randomization tests for XAI evaluation had been introduced by \citet{adebayo2018sanity}, and their shortcomings were discussed by \citet{binder2023shortcomings}.

\textbf{XAI evaluation with residue substitution.} XAI methods can be evaluated, e.g. by replacing those input features that have been identified as most relevant for the classification decision of the model with default values or random noise, and by observing the resulting performance drop. In image recognition, this approach has been demonstrated in the form of ``pixel flipping'' for black and white images, or in the form of pixel substitution with a grey or other color value \citep{bach2015pixel}. In protein function prediction, the residues in each sequence could be sorted according to the attributed relevance. The top $n$ relevant residues could then be substituted with alanine \citep[cf. alanine scanning mutagenesis; ][]{cunningham1989high}, while observing the model performance drop in comparison to the substitution of $n$ random residues. We conducted a residue substitution experiment for EC50 L1, where we substituted an increasing number of relevant or random residues with alanine. Random test protein generation was repeated ten times for establishing the random baseline. (The multi-class problem of EC classification on level L1 might lend itself better to this kind of analysis, because there are only six mutually exclusive classes. For the GO classification case, it must be considered that relevance is computed only for one selected class out of the numerous GO terms, while the performance is calculated over all available classes of this multi-label problem.) \Cref{tab:substitution} shows the result of this residue substitution experiment. The model is highly accurate on the original test split. The performance of ProtBert decreases when more and more relevant as well as random residues are substituted with alanine. Performance decays faster when the most relevant residues are substituted, in comparison to a random replacement. Thus, the residues identified with IG are indeed relevant for the accurate classification decision of the model.

\begin{table}
\caption{
Accuracy of ProtBert on EC50 L1 for an increasing number of substituted residues. Either the $n$ most relevant or $n$ random residues of the proteins from the test split were replaced by alanine. Then, the accuracy of the model trained on the original training data was evaluated on these modified test proteins. Performance drops faster when relevant residues are substituted.}
\label{tab:substitution}
\centering
\begin{tabular}{ccc}
\toprule
$n$ & relevant residues & random residues \\
\midrule
  0 & 0.9898 & 0.9898 \\
  1 & 0.9882 & 0.9898 \\
  2 & 0.9875 & 0.9897 \\
  4 & 0.9853 & 0.9895 \\
  8 & 0.9828 & 0.9895 \\
 16 & 0.9781 & 0.9881 \\
 32 & 0.9525 & 0.9809 \\
 64 & 0.8677 & 0.9296 \\
\bottomrule
\end{tabular}
\end{table}

\textbf{Additional XAI results} for EC and GO, discussed earlier in \Cref{subsec:explainability_analysis_peeking}, are shown in Supplementary \Cref{fig:ec_embedding-EC4050L12,fig:ec_inside,fig:go_inside_catalytic,fig:go_inside_binding}.

\begin{figure}
\centering
\small
\begin{tabular}{ccc}
& E40 & EC50\\
\begin{turn}{90}\hspace{1.8cm} L2 \end{turn} & \includegraphics[width=0.42\columnwidth]{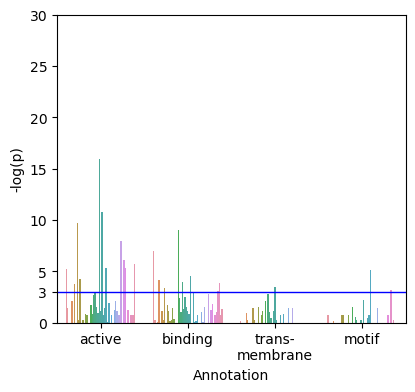} &
\includegraphics[width=0.42\columnwidth]{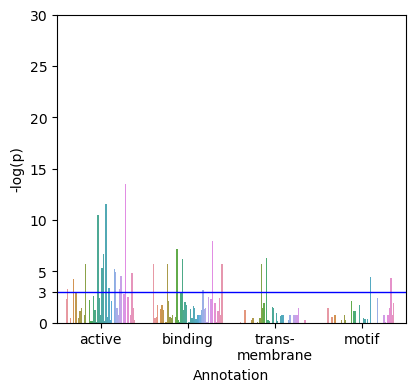}\\
\begin{turn}{90}\hspace{1.8cm} L1 \end{turn}  & \includegraphics[width=0.42\columnwidth]{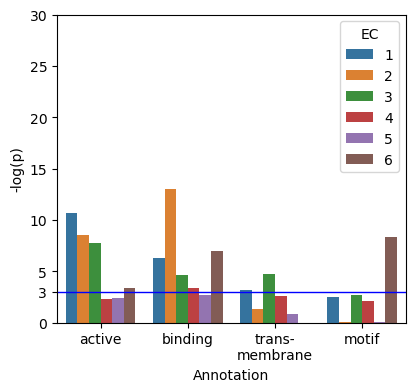} &
\includegraphics[width=0.42\columnwidth]{ec50-L1-annot-rel-corr-embedding-p.png}\\ 
\end{tabular}
\caption{ProtBert embedding layer attributions were correlated with UniProt sequence annotations, for ``EC40'' (left) and ``EC50'' (right), on levels ``L1'' (bottom) and ``L2'' (top).
Significantly positive relevance-annotation-correlation (p$<$0.05, i.e.~above blue line) was observed for ``EC40'' and ``EC50'', on levels ``L1'' and ``L2'', for several enzyme classes (``L1'') and sub-classes (``L2''), and for all annotation types. (Due to the EC hierarchy, more samples were available on level ``L1'' in comparison to ``L2'' for each correlation calculation.)}
\label{fig:ec_embedding-EC4050L12}
\end{figure}

\begin{figure*}
\centering
\small
\begin{tabular}{cccc}
& Corr(relevance, annotation) $>0$ & Relevance $>0$ & Overlay \\
\begin{turn}{90} \hspace{0.5cm} PROSITE\end{turn} & 
\includegraphics[trim={0 0.8cm 0 0}, clip, width=0.16\textwidth]{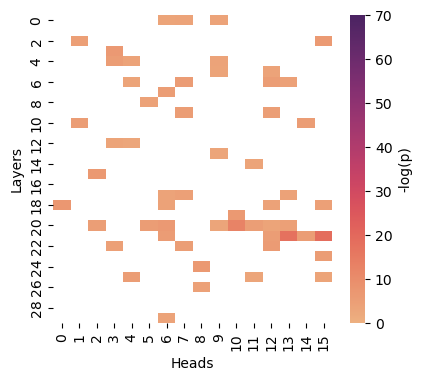} &
\includegraphics[trim={0 0.8cm 0 0}, clip, width=0.16\textwidth]{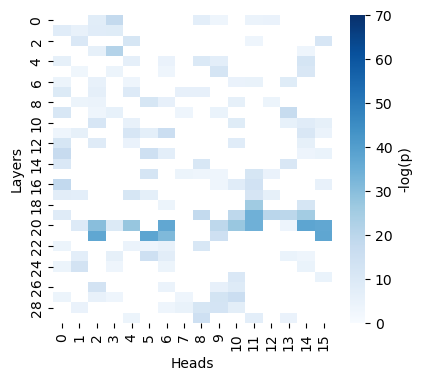} &
\includegraphics[trim={0 0.8cm 0 0}, clip, width=0.16\textwidth]{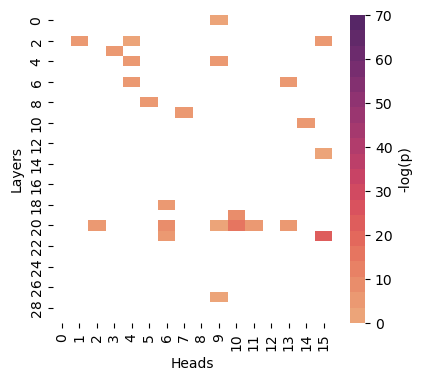} \\
\begin{turn}{90} \hspace{0.5cm} Active sites\end{turn} &
\includegraphics[trim={0 0.8cm 0 0}, clip, width=0.16\textwidth]{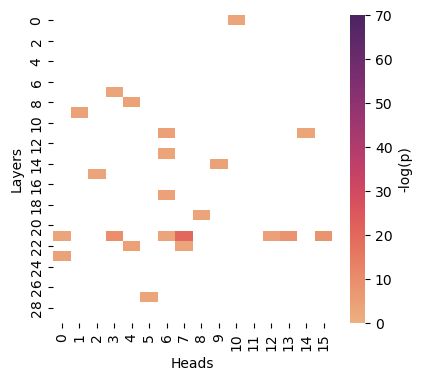} &
\includegraphics[trim={0 0.8cm 0 0}, clip, width=0.16\textwidth]{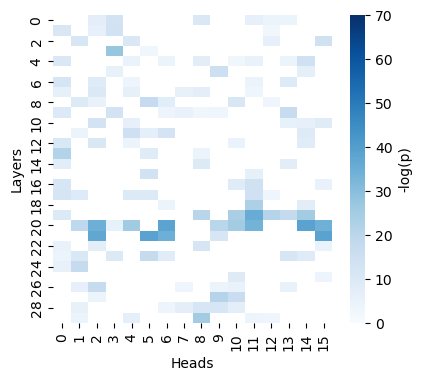} &
\includegraphics[trim={0 0.8cm 0 0}, clip, width=0.16\textwidth]{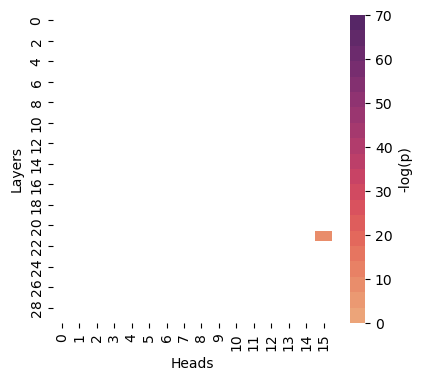} \\
\begin{turn}{90} \hspace{0.5cm} Transmem. reg.\end{turn} &
\includegraphics[width=0.16\textwidth]{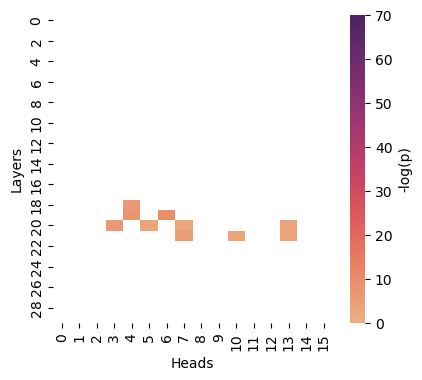} &
\includegraphics[width=0.16\textwidth]{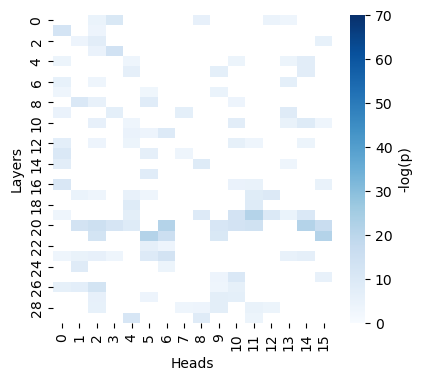} &
\includegraphics[width=0.16\textwidth]{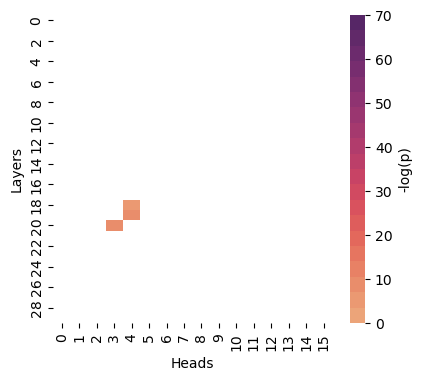} \\
\end{tabular}
\caption{Inside ProtBert; focus on GO ``catalytic activity'' (GO:0003824).
Left panels: Results of the correlation analysis between relevance attributions indicative for the GO term ``catalytic activity'' with PROSITE patterns and UniProt annotations as ``active sites'' and ``transmembrane regions'' per head and layer (negative logarithm of corrected p-values of Wilcoxon signed-rank tests over correlation coefficients; significance thresholds were overlaid as masks). Center panels: (Negative logarithm of corrected p-values of) Wilcoxon tests inspecting whether the relevance (sum along the sequence) was significantly positive. Right panels: Heads with both a significantly positive attribution-annotation-correlation and a significantly positive relevance (overlay of left and center panels). Only results for corresponding PROSITE patterns (top), UniProt ``active sites'' (center) and ``transmembrane regions'' (bottom) are shown. ``Binding sites'' and ``motifs'' are ommited, since a significantly positive attribution-annotation-correlation and combined with a positive relevance was not observed in these cases.}
\label{fig:go_inside_catalytic}
\end{figure*}

\begin{figure*}
\centering
\small
\begin{tabular}{cccc}
& Corr(relevance, annotation) $>0$ & Relevance $>0$ & Overlay \\
\begin{turn}{90} \hspace{0.5cm} PROSITE\end{turn} & 
\includegraphics[trim={0 0.8cm 0 0}, clip, width=0.16\textwidth]{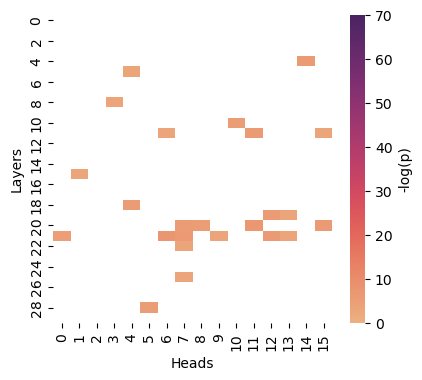} &
\includegraphics[trim={0 0.8cm 0 0}, clip, width=0.16\textwidth]{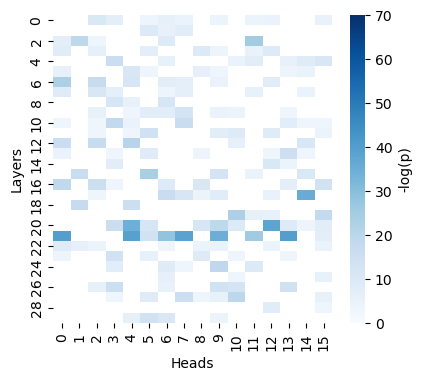} &
\includegraphics[trim={0 0.8cm 0 0}, clip, width=0.16\textwidth]{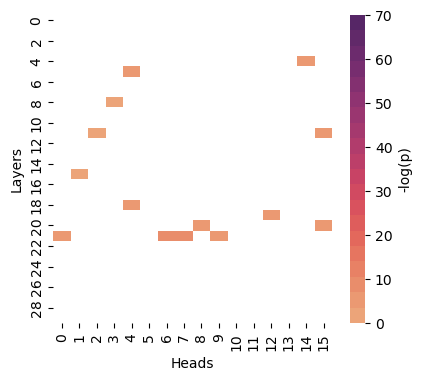} \\
\begin{turn}{90} \hspace{0.5cm} Transmem. reg.\end{turn} &
\includegraphics[width=0.16\textwidth]{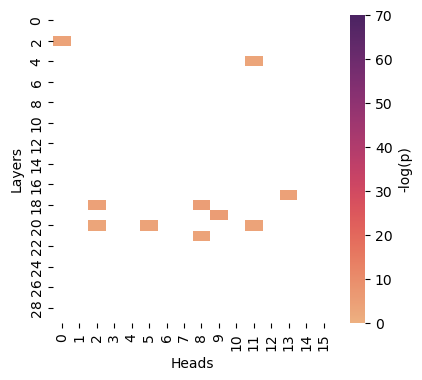} &
\includegraphics[width=0.16\textwidth]{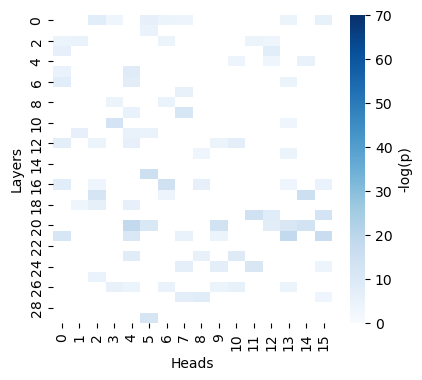} &
\includegraphics[width=0.16\textwidth]{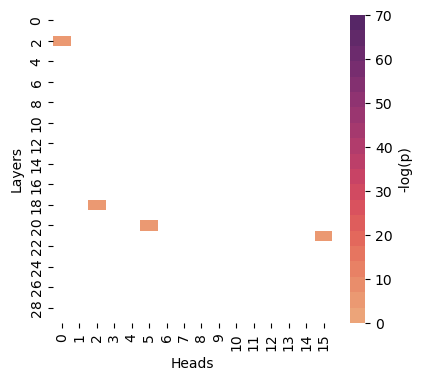} \\
\end{tabular}
\caption{Inside ProtBert; focus on GO ``binding'' (GO:0005488). Left: Correlation of relevance for ``binding'' with corresponding PROSITE patterns (top) or UniProt ``transmembrane regions'' (bottom) per head and layer. Center: Positive relevance. Right: Heads with both a significantly positive attribution-annotation-correlation and a significantly positive relevance.
Results for ``active/binding sites'' and ``motifs'' are not shown, given the absence of significantly positive attribution-annotation-correlations in these cases.}
\label{fig:go_inside_binding}
\end{figure*}

\begin{figure*}
\centering
\begin{tabular}{ccccccc}
& EC1 & EC2 & EC3 & EC4 & EC5 & EC6 \\
\begin{turn}{90}\hspace{0.5cm} \footnotesize{Active sites} \end{turn} & 
\includegraphics[trim={0 0.8cm 0.75cm 0}, clip, height=2.0cm]{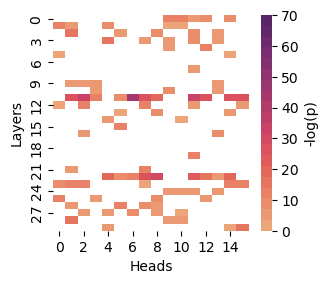} & 
\includegraphics[trim={0.75cm 0.8cm 0.75cm 0}, clip, height=2.0cm]{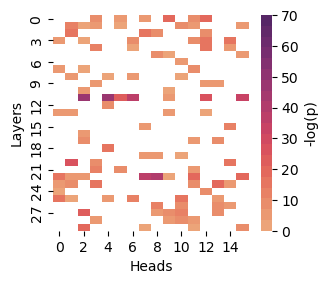} & 
\includegraphics[trim={0.75cm 0.8cm 0.75cm 0}, clip, height=2.0cm]{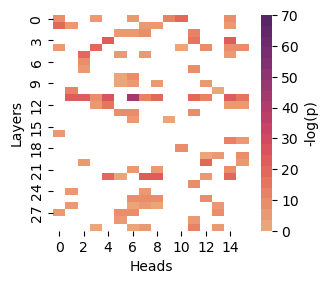} & 
\includegraphics[trim={0.75cm 0.8cm 0.75cm 0}, clip, height=2.0cm]{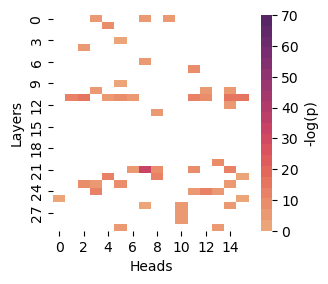} &
\includegraphics[trim={0.75cm 0.8cm 0.75cm 0}, clip, height=2.0cm]{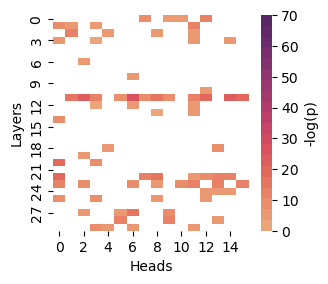} & 
\includegraphics[trim={0.75cm 0.8cm 0 0}, clip, height=2.0cm]{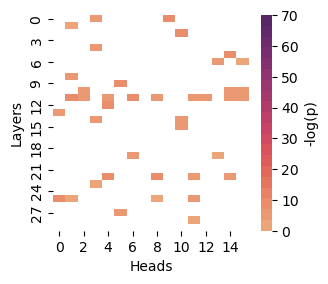} \\
\begin{turn}{90} \hspace{0.5cm} \footnotesize{Binding sites} \end{turn} & 
\includegraphics[trim={0 0.8cm 0.75cm 0}, clip, height=2.0cm]{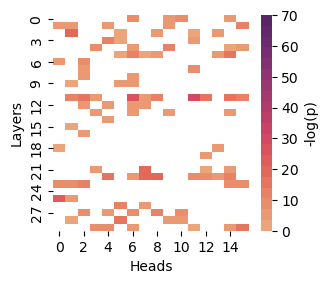} &
\includegraphics[trim={0.75cm 0.8cm 0.75cm 0}, clip, height=2.0cm]{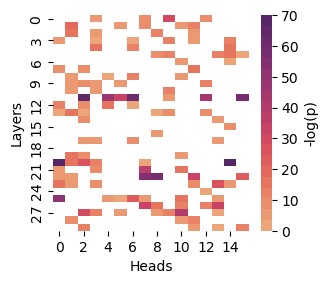} &
\includegraphics[trim={0.75cm 0.8cm 0.75cm 0}, clip, height=2.0cm]{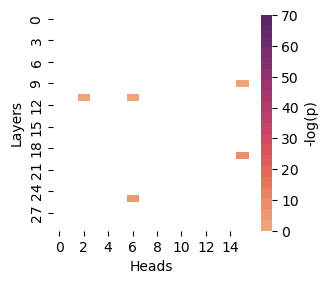} &
\includegraphics[trim={0.75cm 0.8cm 0.75cm 0}, clip, height=2.0cm]{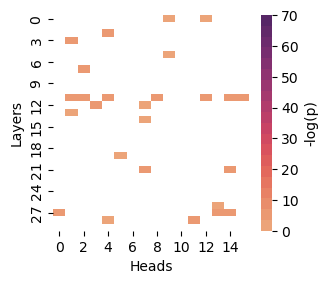} &
\includegraphics[trim={0.75cm 0.8cm 0.75cm 0}, clip, height=2.0cm]{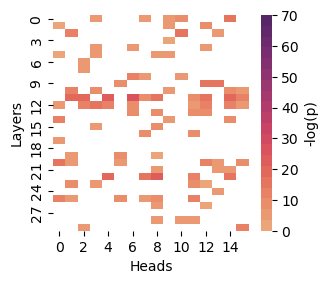} &
\includegraphics[trim={0.75cm 0.8cm 0 0}, clip, height=2.0cm]{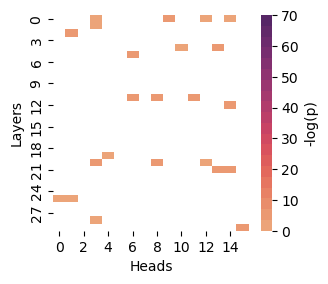} \\
\begin{turn}{90} \hspace{0.3cm} \footnotesize{Transmem. reg.} \end{turn} & 
\includegraphics[trim={0 0.8cm 0.75cm 0}, clip, height=2.0cm]{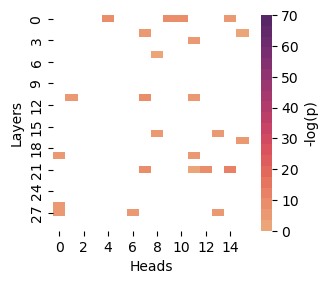} &
\includegraphics[trim={0.75cm 0.8cm 0.75cm 0}, clip, height=2.0cm]{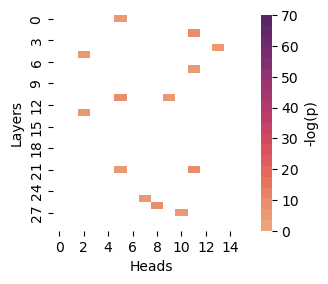} &
\includegraphics[trim={0.75cm 0.8cm 0.75cm 0}, clip, height=2.0cm]{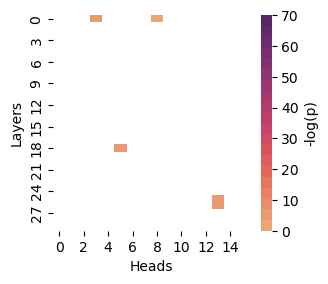} &
\includegraphics[trim={0.75cm 0.8cm 0.75cm 0}, clip, height=2.0cm]{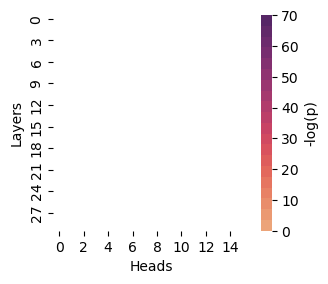} &
\includegraphics[trim={0.75cm 0.8cm 0.75cm 0}, clip, height=2.0cm]{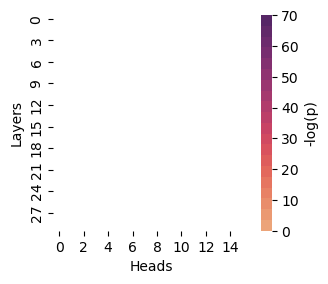} &
\includegraphics[trim={0.75cm 0.8cm 0 0}, clip, height=2.0cm]{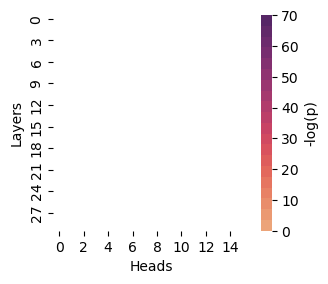} \\
\begin{turn}{90} \hspace{1.0cm} \footnotesize{Motifs} \end{turn} & 
\includegraphics[trim={0 0 0.75cm 0}, clip, height=2.25cm]{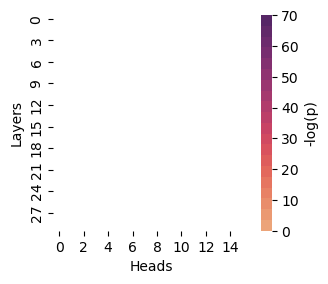} &
\includegraphics[trim={0.75cm 0 0.75cm 0}, clip, height=2.25cm]{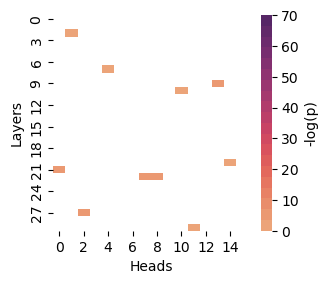} &
\includegraphics[trim={0.75cm 0 0.75cm 0}, clip, height=2.25cm]{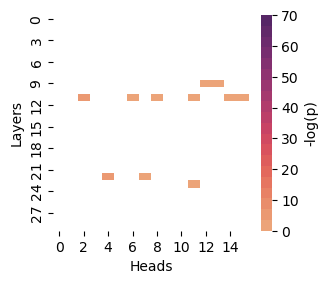} &
\includegraphics[trim={0.75cm 0 0.75cm 0}, clip, height=2.25cm]{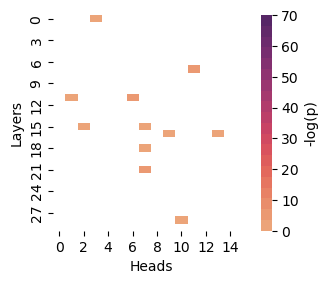} &
\includegraphics[trim={0.75cm 0 0.75cm 0}, clip, height=2.25cm]{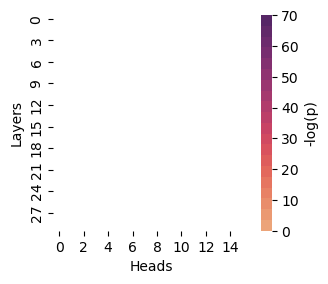} &
\includegraphics[trim={0.75cm 0 0 0}, clip, height=2.25cm]{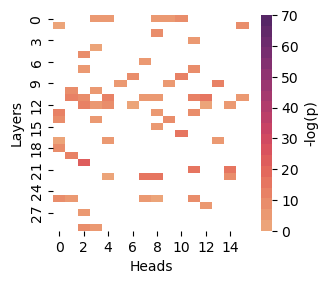} \\
\end{tabular}
\caption{Inside ProtBert finetuned to EC number classification. Rows: Results of the correlation analysis are presented separately for the different types of annotations (UniProt ``active sites'', ``binding sites'', ``transmembrane regions'', ``short sequence motifs''). Columns: Results for the six main EC classes (at level L1). Pixels in the matrix plots represent the 16 transformer heads and 30 layers. Each pixel shows the negative logarithm of the p-value resulting from a Wilcoxon signed-rank test across the coefficients of correlation between relevance attributions and annotation (p$<$0.05; after correction for multiple comparisons and thresholding) overlaid with the mask of those heads/layers that featured a significantly positive relevance too.}
\label{fig:ec_inside}
\end{figure*}

\textbf{Relation between homology and XAI.} Homology describes the relation between proteins that are deemed as originating from a common ancestor, typically because their sequences or structures are sufficiently similar (or, because an intermediary sequence exists that both sequences resemble) \citep{pearson2013introduction}.
To study how homology relates to residue-level relevance attributions of an ML model, homology information must be made available per residue. For example, a consensus sequence can be created via MSA, which can be turned into a binary mask of conserved residues present in the respective protein.
We have correlated attributions with binary masks of short conserved sequence motifs and with \href{https://prosite.expasy.org/prosuser.html}{PROSITE patterns}.
Correlation of attributions with motifs was observed on the embedding level in \Cref{fig:go_inside_membrane}, and, inside of the model for several heads, with motifs and PROSITE patterns in Supplementary \Cref{fig:go_inside_catalytic,fig:go_inside_binding,fig:ec_inside}.

\textbf{Relation between sequence-structure-function and XAI.}
According to the ``sequence-structure-function'' paradigm \citep{koehler2023sequence}, the amino acid sequence determines the secondary protein structure (alpha helices, beta sheets etc.) that determines the tertiary protein structure, i.e. 3D shape, that determines the protein function. An analogy can be drawn to the deep ML models that start with the sequence, build up useful representations over multiple layers, to finally classify the protein function. Here, we aimed at inverting this information flow from sequence to function, and attributed relevance for the function prediction model back to the individual residues in the sequences. Interestingly, we observed a high correlation of attributions w.r.t. the GO term ``membrane'' with annotations as alpha-helical and beta-barrel transmembrane regions (see \Cref{fig:go_embedding}, \Cref{fig:go_inside_membrane}, and \Cref{tab:uniprot_go}). Thus, structurally similar sequences apparently resulted in comparable attributions in this case.

\textbf{Relation to probing and in-silico mutagenesis.} The proposed method can provide a complementary view to probing approaches \citep[][]{vig2021bertology} as well as to in-silico mutagenesis \citep[`ISM'; ][]{bromberg2008comprehensive, raimondi2018large}. ISM and XAI methods have in common that both aim at identifying individual residues that are important for the protein function.
ISM uses ML to mimic alanine scanning mutagenesis \citep{cunningham1989high}, where single residues are replaced with alanine or other amino acids. Then, the (potentially deleterious) effect of the substitution on the protein function is measured \cite[e.g. binding energy change; see][]{bromberg2008comprehensive}, respectively predicted with ML. Thereby, residues can be identified that are essential for the protein function (or stability).
On the other hand, local XAI methods aim at identifying features (here: residues) that are relevant for the decision of a (protein function prediction) model. XAI post-hoc analyses fall into few general categories. Some XAI methods, including IG, are based on gradients, while other methods are ``simulating feature removal'' \citep{covert2021explaining}, e.g. LIME \citep{ribeiro2016why}. In some ways, ISM might be comparable (using the alanine scanning analogy) to feature removal methods where input features are replaced with a given value (alanine) -- in principle, PredDiff \citep{blucher2022preddiff} with a constant imputer.
Whether ISM or XAI methods are better suited in the context of protein function prediction is hard to say. Both take a different perspective, with ISM potentially placing particular emphasis on the protein stability, and with XAI viewing the problem more indirectly through the ``spectacles'' of the function prediction model. Finally, a potential shared limitation of ISM and XAI is that both aim merely at identifying individual residues, potentially neglecting their interaction.

\begin{figure}
\centering
\includegraphics[width=0.6\columnwidth]{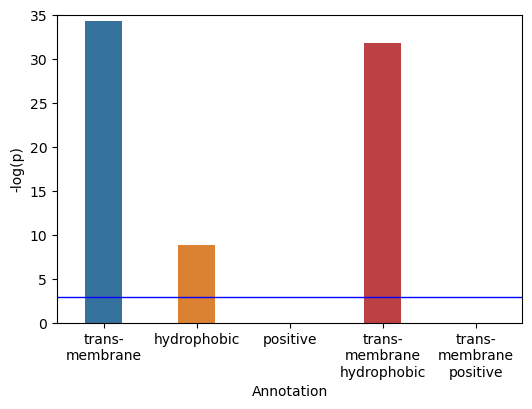} 
\caption{Relevance attributions to residues of membrane proteins correlate (p$<$0.05; above blue line) with binary masks of hydrophobic residues, in particular if they are located in a transmembrane region, but not with positively charged residues. Thus, the model seems to pay particular attention to hydrophobic residues buried in the membrane when inferring whether the GO term ``membrane'' applies to the protein.}
\label{fig:hydrophobicity_charge}
\end{figure}

\textbf{Transmembrane regions, hydrophobicity and charge.} For a more fine-grained inspection of the case of transmembrane regions presented in \Cref{fig:go_embedding}, we additionally tested if the relevance attributions (to residues of proteins that had been labeled with the GO term ``membrane'' and annotated with transmembrane regions) correlated with hydrophobic and, respectively, positively charged residues. Hydrophobic amino acids of transmembrane proteins tend to be located in the hydrophobic core of the membrane, while positively charged residues tend to be attracted by the interface of the membrane with the cytoplasma \citep[positive-inside rule; see ][]{vonheijne1989control, elazar2016interplay, baker2017charged}.
Interestingly, a significant correlation of attributions with residues that were hydrophobic was observed (in particular, if they are situated within a transmembrane region), but not with positively charged residues (see \Cref{fig:hydrophobicity_charge}). Apparently, attention of the model, when aiming at inferring if the GO term ``membrane'' applies to the protein, is particularly drawn to hydrophobic residues buried in the membrane.

\bibliographystyle{natbib}
\small{\bibliography{bibfile}}

\end{document}